\RequirePackage{xr-hyper}
\documentclass[10pt,twocolumn,letterpaper]{article}

\usepackage[pagenumbers]{cvpr} 
\usepackage{booktabs}
\usepackage{multirow}
\usepackage{bm}
\usepackage{amsmath}
\usepackage{amsfonts}
\usepackage{amssymb}
\usepackage{wrapfig}
\usepackage{url}
\usepackage[numbers]{natbib}
\usepackage[resetlabels]{multibib}
\newcites{fir}{Reference}
\newcites{sec}{Reference}

\definecolor{cvprblue}{rgb}{0.21,0.49,0.74}
\usepackage[pagebackref,breaklinks,colorlinks,citecolor=cvprblue]{hyperref}

\def\paperID{2189} 
\def\confName{CVPR}
\def\confYear{2025}

\title{3D Student Splatting and Scooping}

\author{Jialin Zhu$^1$, Jiangbei Yue$^2$, Feixiang He$^1$, He Wang $^{1,3}$\thanks{Corresponding author, he\_wang@ucl.ac.uk}\\
$^1$University College London, UK\ \ \ \ $^2$University of Leeds, UK\\ $^3$AI Centre, University College London, UK}

\newcommand{\sssNameLong}{Student Splatting and Scooping\xspace}
\newcommand{\sssName}{SSS\xspace}

\begin{document}
\maketitle
\begin{abstract}
Recently, 3D Gaussian Splatting (3DGS) provides a new framework for novel view synthesis, and has spiked a new wave of research in neural rendering and related applications. As 3DGS is becoming a foundational component of many models, any improvement on 3DGS itself can bring huge benefits. To this end, we aim to improve the fundamental paradigm and formulation of 3DGS. We argue that as an unnormalized mixture model, it needs to be neither Gaussians nor splatting. We subsequently propose a new mixture model consisting of flexible Student's t distributions, with both positive (splatting) and negative (scooping) densities. We name our model \sssNameLong, or \sssName. When providing better expressivity, \sssName also poses new challenges in learning. Therefore, we also propose a new principled sampling approach for optimization. Through exhaustive evaluation and comparison, across multiple datasets, settings, and metrics, we demonstrate that \sssName outperforms existing methods in terms of quality and parameter efficiency, \eg achieving matching or better quality with similar numbers of components, and obtaining comparable results while reducing the component number by as much as 82\%. 
\end{abstract}

\section{Introduction}
\label{sec:intro}

Presented initially as a neural rendering technique, 3D Gaussian Splatting (3DGS)~\cite{kerbl20233d} has quickly become a versatile component in various systems, \eg geometry reconstruction, autonomous driving~\cite{chen2024survey, fei20243d}. Given its importance as a foundational component, very recently researchers start to investigate possible alternatives to the basic framework of 3DGS, \eg more expressive distributions instead of Gaussians~\cite{hamdi2024ges, li20243d}, more principled optimization~\cite{kheradmand20243d}, which all focus on improving the model expressivity. Our research is among these attempts.

The key to 3DGS' success lies in its name: Gaussian and splatting. 3DGS can be seen as a (unnormalized) Gaussian mixture, which provides two advantages. As a general-purpose distribution, Gaussians can approximate an arbitrary density function hence good expressivity. Also, Gaussians have analytical forms under \eg affine transformation, enabling quick evaluation in 3D-2D projection, thus quick learning from images. Meanwhile, splatting provides a flexible way of identifying only the relevant Gaussians to an image for learning. Despite the success, the framework can still suffer from insufficient expressivity~\cite{kulhanek2024wildgaussians, wu2024implicit}, and low parameter efficiency, \ie needing a large number of components~\cite{sun2024f, lee2024compact}. Therefore, we re-examine the three key components in 3DGS: Gaussian, splatting, and the optimization. Since its underlying principle is essentially to fit a 3D mixture model to a radiance field, we argue it needs not to be restricted to Gaussians or splatting. 

To this end, we propose a simple yet effective generalization of 3DGS. We first replace Gaussians with Student's t distribution with one degree of freedom, referred to simply as t-distribution. Similar to Gaussians, t-distribution also enjoys good properties such as analytical forms under affine transformation. More importantly, t-distribution can be seen as a generalization of Gaussians and therefore is more expressive. t-distribution has a control parameter for the tail fatness, representing distributions ranging from Cauchy distribution to Gaussian distribution, and any distribution inbetween. Compared with Gaussians, Cauchy is fat-tailed, \ie a single Cauchy can cover a larger area with comparatively higher densities than a Gaussian. Furthermore, by making this control parameter learnable, we learn components with a wide range of varying tail thicknesses. 

Next, we extend the splatting scheme which only operates in the positive density space. Inspired by mixture models with negative components~\cite{loconte2024subtractive}, we propose to employ both positive and negative components to splat (adding) and scoop (subtracting) densities from the model. This leads to more complex mathematical forms than 3DGS but we derive their close-form gradients for learning. 

Finally, as the increased model complexity, optimization methods based on naive stochastic gradient descent become insufficient, due to parameter coupling. Therefore, we propose a principled sampling scheme based on Stochastic Gradient Hamiltonian Monte Carlo (SGHMC). 

We refer to our model as \sssNameLong (\sssName). \sssName is evaluated on multiple datasets and compared with existing methods. Experiments show that \sssName achieves higher quality often with fewer number of components, demonstrating more expressivity and higher parameter efficiency. Formally, our contributions include:
\begin{itemize}
    \item A new model named \sssNameLong (\sssName), which is highly expressive and parameter efficient.
    \item A new mixture model with flexible components learned from a set of distribution families for neural rendering.
    \item A mixture model with negative components, which extends the learning into the negative density space.
    \item A principled sampling approach to tackle parameter coupling during learning.
\end{itemize}

\section{Related Work}
\label{sec:related_work}
\paragraph{3D Reconstruction and Novel View Synthesis}
3D reconstruction and Novel View Synthesis have been long-standing research topics in computer vision.
Traditional methods mainly include Multi-View Stereo (MVS)~\cite{seitz2006comparison} and Structure from Motion (SFM)~\cite{ullman1979interpretation}. 
Recently, the advent of Deep Learning has brought important changes to the field. In particular, the techniques based on Neural Radiance Field (NeRF)~\cite{mildenhall2021nerf} and 3D Gaussian Splatting (3DGS)~\cite{kerbl20233d} have set new state-of-the-art (SOTA) benchmarks.

\paragraph{NeRF methods}
NeRF~\cite{mildenhall2021nerf} proposes to implicitly encode the radiance field of a 3D object/scene into a neural network and renders the 3D geometry and textures through a continuous volume rendering function. Since then, a large number of methods based on NeRF have been proposed, namely NeRF++~\cite{zhang2020nerf++}, Mip-NeRF~\cite{barron2022mip} and Mip-NeRF360~\cite{barron2022mip} to improve rendering quality, Plenoxels~\cite{fridovich2022plenoxels} and Instant-NGP~\cite{muller2022instant} to accelerate NeRF training, D-NeRF~\cite{pumarola2021d} to extend NeRFs to dynamic scenes, DreamFusion~\cite{pooledreamfusion} and Zero-1-to-3~\cite{liu2023zero} to employ it for text-to-3d generation models, etc. However, the biggest drawback of NeRF is that the ray casting process for rendering is time consuming. Despite the effort in improving its rendering efficiency \eg SNeRG~\cite{hedman2021baking} and mobileNeRF~\cite{chen2023mobilenerf}, it still cannot be used for real-time rendering in most cases.

\paragraph{Splatting methods}
3DGS~\cite{kerbl20233d} solves the above problem for real-time rendering, by replacing the volume rendering with a differentiable rasterization method, which achieves the SOTA render quality. 
3DGS uses 3D Gaussian as the primitive for the splatting method~\cite{zwicker2001ewa, zwicker2002ewa}. It directly projects 3D Gaussians onto the 2D image plane through view/projective transformation for rasterization.
Similar to NeRFs, prolific follow up research has been conducted based on 3DGS. GS++~\cite{huang2024gs++} and Mip-Splatting~\cite{yu2024mip} aim to improve rendering quality, 4D Gaussian Splatting~\cite{wu20244d} and Deformable 3D Gaussians~\cite{yang2024deformable} extend 3DGS to dynamic scenes, Dreamgaussian~\cite{tangdreamgaussian} employs 3DGS for text-to-3D tasks. One particular line of research is to improve the fundamental paradigm of 3DGS. This includes FreGS~\cite{zhang2024fregs}, 3DGS-MCMC~\cite{kheradmand20243d} and Bul{\`o} et al.~\cite{rota2024revising} which optimize the training process and adaptive density control in 3DGS,  Scaffold-GS~\cite{lu2024scaffold} and Implicit Gaussian Splatting~\cite{wu2024implicit} which combine grid representation with 3DGS for better rendering quality. More recently, there is also research exploring different primitives other than 3D Gaussians. 2DGS~\cite{huang20242d} obtains better surface reconstruction by changing the primitives from 3D Gaussian to 2D Gaussian for aligning the 3D scene. GES~\cite{hamdi2024ges} uses a generalized exponential kernel to increase the expression ability of primitives and reduce memory cost. 3DHGS~\cite{li20243d} decomposes one Gaussian into two half-Gaussians to obtain asymmetry and better expressivity. 

Our research is among the few recent efforts in improving the fundamental formulation of 3DGS. Different from them, we propose to use more expressive and flexible distributions, 3D Student's t distribution, as the basic primitive. In addition, we also use both positive and negative densities to extend the optimization into the negative density space for better representation. Finally, we propose a principled sampling approach for learning, deviating from most of the above research.

\section{Methodology}
\label{sec:methodology}

\subsection{Preliminaries: 3DGS as a mixture model}
\label{sec:3_1}
3DGS essentially fits a (unnormalized) 3D Gaussian mixture model to a radiance field~\cite{kerbl20233d}:
\begin{equation}
    P(x) = \sum w_i G_i(x), \text{ }G(x) = e^{-\frac{1}{2}(x-\mu)^T\Sigma^{-1}(x-\mu)}
\end{equation}
where $w_i > 0$. $\mu$ is the center position of Gaussian. $\Sigma \in \mathbb{R}^{3\times3}$ is the covariance matrix, parameterized by a scaling matrix $S$ and a rotation matrix $R$ to maintain its positive semi-definiteness: $\Sigma = RSS^TR^T$. Additionally, every 3D Gaussian is associated with opacity $o \in [0,1]$ and color $c\in \mathbb{R}^{27}$ which is represented by spherical harmonics and view-dependent~\cite{kerbl20233d}.$w_i$ is determined by $o$, $c$ and compositing values after projection.

Since the mixture can only be evaluated in 2D, when rendering an image, a 3D Gaussian is projected onto the 2D image plane, $G^{2D}$, via integrating it in the camera view space along a ray, for computing a pixel color:
\begin{equation}
    C(u) = \sum_{i=1}^{N}c_{i}o_{i}G^{2D}_{i}(u)\prod_{j=1}^{i-1}(1-o_{j}G^{2D}_{j}(u)).
    \label{eq:GS_color}
\end{equation}
where $N$ is the number of the Gaussians that intersect with the ray cast from the pixel $u$. Finally, the Gaussian parameters, opacity, and colors are learned based on the observed 2D images. \cref{eq:GS_color} reveals that $C(u)$ can be seen as a 2D Gaussian mixture, except that now a component weight is also a function of other components, introducing additional cross-component interactions.

Numerically, Gaussians, as the mixture component, are closed under affine transformation and marginalization of variables, so that the forward/backward pass can be quickly computed. 3DGS is a \textit{monotonic} mixture as it is additive, \ie $w_i>0$. Due to the success of 3DGS, existing works have since followed this paradigm~\cite{zhang2024fregs, lu2024scaffold, huang2024gs++, yu2024mip}.

\subsection{Student's t as a basic component}
\label{sec:3_2}
We propose an unnormalized t-distribution mixture model, where a t-distribution is defined by a mean (location) $\mu\in\mathbb{R}^3$, a covariance matrix (shape) $\Sigma=RSS^TR^T\in\mathbb{R}^{3\times3}$, a degree of freedom (tail-fatness) $\nu\in\left[1, +\infty\right)$, associated with opacity $o$, and color $c$:
\begin{align}
    &P(x) = \sum w_i T_i(x) \text{ , } w_i > 0\nonumber \\
    &T(x|\nu)= [1+\frac{1}{\nu}(x-\mu)^T\Sigma^{-1}(x-\mu)]^{-\frac{\nu+3}{2}},
    \label{eq:t-dist}
\end{align}
where we can drop the scalar $\frac{\Gamma(\nu+3)/2}{\Gamma(\nu/2)\nu^{\frac{3}{2}}\pi^{\frac{3}{2}}|\Sigma|^{\frac{1}{2}}}$ in the original t-distribution safely to facilitate learning. 
\begin{figure}[tb]
    \centering
    \includegraphics[width=0.8\columnwidth]{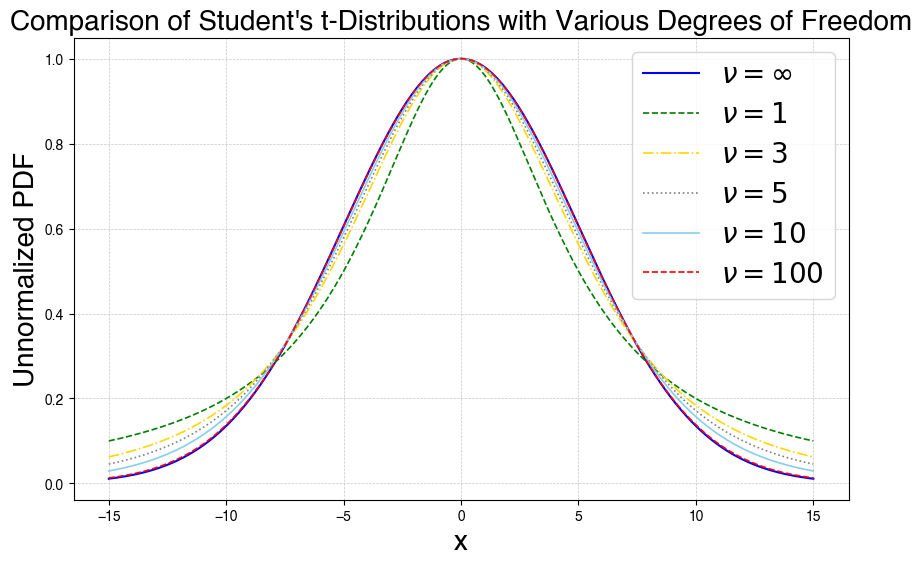}
    \caption{Student's t with varying degrees of freedom $\nu$. (standard deviation is 5).}
    \label{fig:student_t}
\end{figure}
The choice is driven by two main factors. First, t-distribution is a strong generalization of Gaussians. As shown in \cref{fig:student_t}, when $\nu\rightarrow1$, $T\rightarrow Cauchy$; when $\nu\rightarrow \infty$, $T\rightarrow Gaussian$. So t-distribution can capture what Gaussians capture and beyond. Furthermore, since Cauchy is fat-tailed, it can cover larger areas with higher densities than Gaussians therefore potentially reducing the number of components. As $\nu$, $\mu$ and $\Sigma$ are learnable, \sssName becomes a mixture of components learned from an infinite number of distribution families, instead of one family~\cite{kerbl20233d}, providing further flexibility.

The second reason for t-distribution is it also provides good properties similar to Gaussians, \eg close under affine transformation and marginalization of variables. Rendering a pixel requires an affine transformation, then a projective transformation, followed by an integration along a ray, to be applied to a component, which has a simple form in 3DGS. In \sssName, t-distribution also has a close form: 
\begin{align}
    T^{2D}(u) = &[1+\frac{1}{\nu}(u-\mu^{2D})^T(\Sigma^{2D})^{-1}(u-\mu^{2D})]^{-\frac{\nu+2}{2}} \nonumber \\
    \mu^{2D} &= (W\mu+t)_{1:2}/((W\mu+t)_{3}) \nonumber\\
    \Sigma^{2D} &= (JW \Sigma W^TJ^T)_{1:2,1:2},
    \label{eq:t_2D}
\end{align}
where the subscripts select the corresponding rows and columns. $W$, $t$ and $J$ are the affine transformation (\ie scale, translation) and (approximated) projective transformation~\cite{zwicker2001ewa, zwicker2002ewa}. This enables us to easily derive the key gradients for learning shown in the supplementary material (SM), unlike existing research also using alternative mixture components but requiring approximation~\cite{hamdi2024ges}.

In summary, the mixture of learnable t-distributions enhances the representational power and provides good mathematical properties for learning.

\subsection{Splatting and Scooping}
\label{sec:3_3}

\begin{figure}
    \centering
    \def\svgwidth{1.0\linewidth}
    \fontsize{7pt}{9pt}\selectfont
\begingroup%
  \makeatletter%
  \providecommand\color[2][]{%
    \errmessage{(Inkscape) Color is used for the text in Inkscape, but the package 'color.sty' is not loaded}%
    \renewcommand\color[2][]{}%
  }%
  \providecommand\transparent[1]{%
    \errmessage{(Inkscape) Transparency is used (non-zero) for the text in Inkscape, but the package 'transparent.sty' is not loaded}%
    \renewcommand\transparent[1]{}%
  }%
  \providecommand\rotatebox[2]{#2}%
  \newcommand*\fsize{\dimexpr\f@size pt\relax}%
  \newcommand*\lineheight[1]{\fontsize{\fsize}{#1\fsize}\selectfont}%
  \ifx\svgwidth\undefined%
    \setlength{\unitlength}{2594.88782988bp}%
    \ifx\svgscale\undefined%
      \relax%
    \else%
      \setlength{\unitlength}{\unitlength * \real{\svgscale}}%
    \fi%
  \else%
    \setlength{\unitlength}{\svgwidth}%
  \fi%
  \global\let\svgwidth\undefined%
  \global\let\svgscale\undefined%
  \makeatother%
  \begin{picture}(1,0.27709126)%
    \lineheight{1}%
    \setlength\tabcolsep{0pt}%
    \put(0,0){\includegraphics[width=\unitlength,page=1]{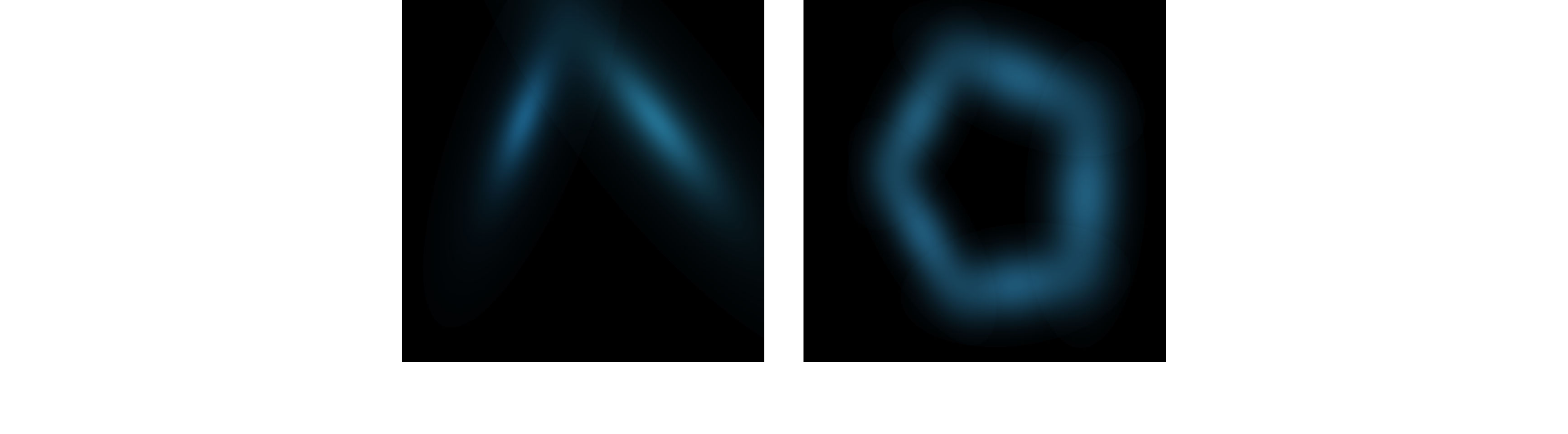}}%
    \put(0.1147016,0.01683931){\makebox(0,0)[t]{\lineheight{1.25}\smash{\begin{tabular}[t]{c}(a) Topology\\data\end{tabular}}}}%
    \put(0.37187064,0.01683931){\makebox(0,0)[t]{\lineheight{1.25}\smash{\begin{tabular}[t]{c}(b) 2 positive\\components\end{tabular}}}}%
    \put(0.62721905,0.01683931){\makebox(0,0)[t]{\lineheight{1.25}\smash{\begin{tabular}[t]{c}(c) 5 positive\\components\end{tabular}}}}%
    \put(0.88438809,0.01683931){\makebox(0,0)[t]{\lineheight{1.25}\smash{\begin{tabular}[t]{c}(d) 1 positive and\\1 negative components\end{tabular}}}}%
    \put(0,0){\includegraphics[width=\unitlength,page=2]{figures/negative_lowres.pdf}}%
  \end{picture}%
\endgroup%

    \caption{\textbf{High parameter efficiency by negative components}. We use a torus with only ambient lighting and frontal views (a), where the challenge is to capture the shape topology with as few components as possible. We initialize the component means near the center. Only using positive densities either underfits if two components are used (b), or requires at least 5 components to capture the topology correctly (c). In contrast, in (d), we only need two components (one positive and one negative), to capture the topology of the shape. Both components are co-located at the center of the torus. The positive component covers the torus but also the hole, while the negative component subtracts densities in the middle to make a hole.}
    \label{fig:negative}
\end{figure}

While monotonic mixture models are powerful, a \textit{non-monotonic} mixture model recently has been proposed by introducing negative components~\cite{loconte2024subtractive}, arguing that it is sub-optimal to only operate in the positive density space:
\begin{equation}
P(x) = (\sum w_i T_i(x))^2=\sum_{i=1}^K \sum_{j=1}^K w_i w_j T_i(x) T_j(x)
\label{eq:nmmm}
\end{equation}
where $w\in \mathbb{R}$. In our problem, a negative density makes good sense as it can be seen as subtracting a color. However, our experiments using \cref{eq:nmmm} show that it is not ideal as it introduces interactions between every pair of components, increasing the model evaluation complexity to $O(n^2)$ where $n$ is the number of components in the model, making it significantly slower than before. Therefore, we still use \cref{eq:t-dist} but with $w\in\mathbb{R}$ instead of $w>0$ where $w_i = c_{i}o_{i}\prod_{k=1}^{i-1}(1-o_{k}T^{2D}_{k}(u))$ and $o \in [-1,1]$. Normally this might cause issues as \cref{eq:t-dist} is then not well defined with negative components. However, we can parameterize the density in an energy-based form explained later which is well defined. In learning, we constrain the opacity by a $tanh$ function so that positive and negative components can dynamically change signs while being bounded. Introducing negative t-distribution can enhance the representation power and the parameter efficiency. We show a simple experiment in \cref{fig:negative}, where fewer components are needed to fit the shape topology of a torus. In \sssName, a component with negative densities is equivalent to removing its color from the mixture. Negative components are particularly useful in subtracting colors.

\subsection{Learning via sampling}
\label{sec:3_4}
Recently, it is argued that principled sampling is better in 3DGS, \eg Markov Chain Monte Carlo~\cite{kheradmand20243d}, instead of naive stochastic gradient descent (SGD).
Empirically, we found training \sssName involves learning more tightly coupled parameters compared with 3DGS, namely among $\nu$, $\mu$, and $\Sigma$. We speculate that this is because changing $\nu$ in learning is changing the family of distributions within which we optimize $\mu$ and $\Sigma$. Therefore, we propose a sampling scheme that mitigates such coupling, based on Stochastic Gradient Hamiltonian Monte Carlo (SGHMC). 

Starting from the Hamiltonian Monte Carlo, we first parameterize the posterior distribution as:
\begin{equation}
    P(\theta, r) \propto exp(-L_{\theta}(x) -\frac{1}{2}r^TIr)
    \label{eq:posterior}
\end{equation}
where $L_{\theta}(x)$ is our loss function, $I$ is an identity matrix, $r$ is a momentum auxiliary variable, and $\theta$ is the learnable parameters. 
This is because \cref{eq:t-dist} with $w\in\mathbb{R}$ is not a well-defined distribution, which makes direct sampling difficult. Using an energy function circumvents this issue and prescribes the high density regions of good $\theta$. Intuitively, we would like to sample $\theta$ to minimize $L_{\theta}(x)$. In addition, to decouple parameters during learning, the momentum term $\frac{1}{2}r^TIr$ creates frictions for each dimension of the parameter space, enabling adaptive learning for each parameter.

For $L_{\theta}(x)$, our rendering function computes the pixel value based on the $N$ components associated with a ray:
\begin{equation}
    C(u) = \sum_{i=1}^{N}c_i o_i T^{2D}_{i}(u)\prod_{j=1}^{i-1}(1-o_jT^{2D}_{j}(u)).
    \label{eq:SSS_color}
\end{equation}
where $u$ is the pixel. $c$ and $o$ are the color and opacity associated with a component $T$. We then employ the following loss function~\cite{kheradmand20243d}:
\begin{align}
L = &(1-\epsilon_{D-SSIM}) L_1 + \epsilon_{D-SSIM} L_{D-SSIM} \nonumber \\
&+ \epsilon_o \sum_i |o_i|_1 + \epsilon_{\Sigma}\sum_i\sum_j |\sqrt{\lambda_{i,j}}|_1
\end{align}
where the $L_1$ norm, and the structural similarity $L_{D-SSIM}$ loss aim to reconstruct images, while the last two terms act as regularization, with $\lambda$ being the eigenvalues of $\Sigma$. The regularization applied to the opacity ensures that the opacity is big only when a component is absolutely needed. The regularization on $\lambda$ ensures the model uses components as spiky as possible (i.e. small variances). Together, they minimize the needed number of components~\cite{kheradmand20243d}.

Furthermore, directly sampling from \cref{eq:posterior} requires the full gradient of $U=L_{\theta}(x) -\frac{1}{2}r^TIr$ which is not possible given the large number of training samples. Therefore, replacing the full gradient with stochastic gradient will introduce a noise term: $\nabla \hat{U} = \nabla U + \mathcal{N}(0, V)$, where $\mathcal{N}$ is Normal and $V$ is the covariance of the stochastic gradient noise. Under mild assumptions~\cite{chen2014stochastic}, sampling \cref{eq:posterior} using stochastic gradients becomes (with detailed derivation in the SM):
\begin{align}
    d\theta = &M^{-1}rdt \nonumber \\
    dr = &-\nabla U(\theta)dt - CM^{-1}rdt + \mathcal{N}(0, 2Cdt)
\label{eq:update}
\end{align}
where $\mathcal{N}$ is Gaussian noise, $M$ is a mass matrix, and $C$ is a control parameter dictating the friction term $CM^{-1}rdt$ and noise $\mathcal{N}(0, 2Cdt)$. In our problem, it is crucial to design good friction and noise scheduling. The effect of this principled sampling method is further discussed in SM.

\subsubsection{Friction and Noise Scheduling}
We first use SGHMC on $\mu$ and Adam on the other parameters. To learn $\mu$, we modify \cref{eq:update} to:
\begin{align}
    \mu_{t+1} &= \mu_t - \varepsilon^2 \left [\frac{\partial L}{\partial \mu}\right ]_{t} + F + N \nonumber \\
    F &= \sigma(o)\varepsilon(1 - \varepsilon C)r_{t-1} \nonumber \\
    N &= \sigma(o)\mathcal{N}(0, 2\varepsilon^{\frac{3}{2}}C) \nonumber \\
    r_{t+1} &= r_{t} - \varepsilon\left[\frac{\partial L}{\partial \mu}\right]_{t+1} - \varepsilon C r_{t-1} + \mathcal{N}(0, 2\varepsilon C) \nonumber \\
    &\text{where }\sigma(o) = \sigma(-k(o-t))
    \label{eq:sigmoidUpdate}
\end{align}
where $\varepsilon$ is the learning rate and decays during learning. $\mathcal{N}$ is Gaussian noise. $o$ is the opacity. The main difference between \cref{eq:sigmoidUpdate} and \cref{eq:update} is we now have adaptive friction $F$ and noise $N$ for $\mu$. $\sigma$ (sigmoid function) switches on/off the friction and noise. We use $k=100$ and $t=0.995$, so that it only activates for components with opacity lower than 0.005. When it is activated, friction and noise are added to these components. Note that if $F$ is disabled, \cref{eq:sigmoidUpdate} is simplified to a Stochastic Gradient Langevin Dynamics scheme~\cite{welling2011bayesian}. 

When learning, we initialize with a sparse set of (SfM) points without normals~\cite{kerbl20233d}, run \cref{eq:sigmoidUpdate} without $F$ for burn-in for exploration, then run the full sampling for exploitation until convergence. During burn-in stage, we multiply $N$ by the covariance $\Sigma$ of the component to maintain the anisotropy profile of the t-distribution. After the burn-in, $\Sigma$ is removed, and the anisotropy is then maintained by $F$ due to the momentum $r$.

\paragraph{Key gradients} Overall, the key learnable parameters for each component in \sssName include the mean $\mu$, covariance $\Sigma$ (\ie $S$ and $R$), color $c$, opacity $o$, and degree parameter $\nu$. To compute \cref{eq:update}, the key gradients are $\frac{\partial L}{\partial \mu}$, $\frac{\partial L}{\partial S}$ ,$\frac{\partial L}{\partial R}$, $\frac{\partial L}{\partial c}$, $\frac{\partial L}{\partial o}$, and $\frac{\partial L}{\partial \nu}$. For simplicity, we give them in the SM.

\subsubsection{Adding and Recycling Components}
Components can become nearly transparent during sampling, \ie near zero opacity. In 3DGS, they are removed. Recently, it is argued that they should be recycled~\cite{kheradmand20243d}, by relocating them to a high opacity component. However, careful consideration needs to be taken as the overall distribution before and after relocation should be the same~\cite{rota2024revising}. When moving some components to the location of another component, this is ensured by:

\begin{equation}
    \text{min} \int_{-\infty}^{\infty} ||C_{new}(\mu) - C_{old}(\mu) ||^2_2 d\mu
    \label{eq:colorDiff}
\end{equation}
where $C_{new}$ and $C_{old}$ are the color after and before relocation respectively. Minimizing this integral ensures the smallest possible pixel-wise color changes over the whole domain. Minimizing \cref{eq:colorDiff} in \sssName leads to:
\begin{align}
    &\mu_{new} = \mu_{old} \text{ , }(1 - o_{new})^N = (1-o_{old}) \nonumber \\
    &\Sigma_{new} = (o_{old})^2 \frac{\nu_{old}}{\nu_{new}} (\frac{\beta(\frac{1}{2}, \frac{\nu_{old}+2}{2})}{K})^2 \Sigma_{old} \nonumber \\
    &K = \sum_{i=1}^N \sum_{k=0}^{i-1}\begin{pmatrix}i-1\\k\end{pmatrix} (-1)^k(o_{new})^{k+1} Z) \nonumber \\
    &Z = \beta(\frac{1}{2}, \frac{(k+1)(\nu_{new}+3)-1}{2})
    \label{eq:relocation}
\end{align}
$\mu_{new}$ and $\mu_{old}$, $o_{new}$ and $o_{old}$, $\Sigma_{new}$ and $\Sigma_{old}$ are the mean, opacity, and covariance after and before relocation respectively. $N$ is the total number of components after relocation, \ie moving $N$-1 low opacity components to the location of 1 high opacity component. $\beta$ is the beta function. We leave the detailed derivation in the SM. Note we do not distinguish between positive and negative components during relocation. This introduces a perturbation on the sign of the opacity. In \cref{eq:relocation}, if $o_{old} > 0$ then all $o_{new} > 0$, or otherwise $o_{new} < 0$ if $o_{old} < 0$, regardless their original opacity signs. This sign perturbation in practice helps the mixing of the sampling. Furthermore, to ensure the sampling stability, we limit the relocation to a maximum of $5\%$ of the total components at a time. Finally, we also add new components when needed, but do not use the adaptive density control (clone and split) in 3DGS~\cite{kerbl20233d}. Instead, we add 5\% new components with zero opacity and then recycle them.

\section{Experiments}
\label{sec:experiments}

\subsection{Experimental setting}
\paragraph{Datasets and Metrics}
Following existing research, we employ 11 scenes from 3 datasets, including 7 public scenes from Mip-NeRF 360~\cite{barron2022mip}, 2 outdoor scenes from Tanks \& Temples~\cite{knapitsch2017tanks}, and 2 indoor scenes from Deep Blending~\cite{hedman2018deep}. Also, following the previous evaluation metrics, we use Peak Signal-to-Noise Ratio (PSNR), Structural Similarity Index Metric (SSIM)~\cite{wang2004image}, and Learned Perceptual Image Patch Similarity (LPIPS)~\cite{zhang2018unreasonable}. We provide average scores of each dataset, and detailed scores are in the SM. 

\paragraph{Baselines}
Due to there being many publications based on 3DGS, we only choose the original 3DGS~\cite{kerbl20233d} and the most recent work that focuses on improving the fundamental paradigm of 3DGS and has achieved the best performances. The methods include Generalized Exponential Splatting (GES)~\cite{hamdi2024ges} and 3D Half-Gaussian Splatting (3DHGS)~\cite{li20243d} which also use different (positive only) mixture components, Scaffold-GS~\cite{lu2024scaffold} and Fre-GS~\cite{zhang2024fregs} which optimize the training procedure to achieve faster convergence and better results, 3DGS-MCMC~\cite{kheradmand20243d} which proposes a principled MCMC sampling process, and Mip-NeRF~\cite{barron2021mip} which is a state-of-the-art method with Neural Radiance Field (NeRF)~\cite{mildenhall2021nerf}. Overall, our baselines comprehensively include methods with new mixture components, new optimization approaches, and the SOTA quality.

The results of all baselines in general benchmarking are from their papers. In addition, we run their codes with other settings for more comparison. Since not all baseline methods are implemented in exactly the same setting, we need to adapt them for comparison. 
These details are in the SM.

\subsection{General Benchmarks}
\begin{figure*}[tb]
    \centering
    \def\svgwidth{1.0\linewidth}
    \fontsize{8pt}{10pt}\selectfont
    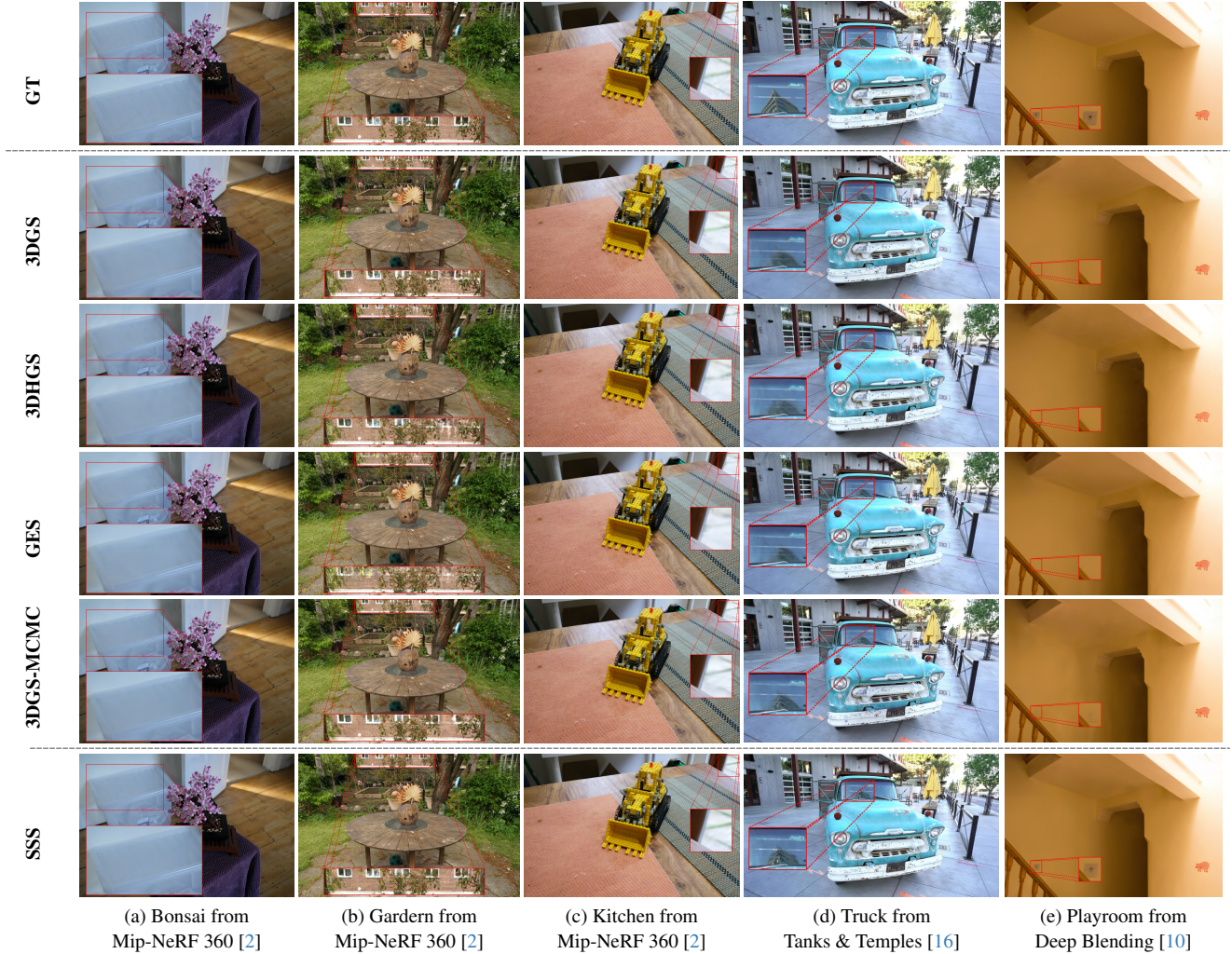
    \caption{\textbf{Visual comparison} Zoom-in for better visualization. (a) \sssName restores better the indentations of the box lid; (b) \sssName is the best at detailing windows in the upper center; (c) Only the image rendered by \sssName contains the green track detail in the upper right corner; (d) \sssName is the best at restoring the reflection in the front window of the truck; (e) \sssName perfectly restores the light switch next to the stairs.}
    \label{fig:full_visual}
\end{figure*}

\begin{table*}[tb]
\begin{tabular}{l|ccc|ccc|ccc}
    Dataset & \multicolumn{3}{c|}{Mip-NeRF360 Dataset}  & \multicolumn{3}{c|}{Tanks\&Temples} & \multicolumn{3}{c}{Deep Blending}\\
    Method|Metric & PSNR $\uparrow$   & SSIM$\uparrow$   & LPIPS$\downarrow$ & PSNR $\uparrow$   & SSIM$\uparrow$   & LPIPS$\downarrow$ & PSNR $\uparrow$   & SSIM$\uparrow$   & LPIPS$\downarrow$ \\
    \hline 
    Mip-NeRF & 29.23 & 0.844 & 0.207 & 22.22 & 0.759 & 0.257 & 29.40 & 0.901 & 0.245\\
    3DGS & 28.69 & 0.870 & \cellcolor{yellow!40}0.182 & 23.14 & 0.841 & 0.183 & 29.41 & 0.903 & \cellcolor{yellow!40}0.243\\
    GES & 26.91 & 0.794 & 0.250 & 23.35 & 0.836 & 0.198 & 29.68 & 0.901 & 0.252\\
    3DHGS & \cellcolor{yellow!40}29.56 & \cellcolor{yellow!40}0.873 & \cellcolor{orange!40}0.178 & \cellcolor{orange!40}24.49 & \cellcolor{yellow!40}0.857 & \cellcolor{orange!40}0.169 & 29.76 & \cellcolor{yellow!40}0.905 & \cellcolor{orange!40}0.242\\
    Fre-GS & 27.85 & 0.826 & 0.209 & 23.96 & 0.841 & 0.183 & \cellcolor{yellow!40}29.93 & 0.904 & \cellcolor{red!40}0.240\\
    Scaffold-GS & 28.84 & 0.848 & 0.220 & 23.96 & 0.853 & \cellcolor{yellow!40}0.177 & \cellcolor{red!40}30.21 & \cellcolor{orange!40}0.906 & 0.254\\
    3DGS-MCMC & \cellcolor{orange!40}29.89 & \cellcolor{red!40}0.900 & 0.190 & \cellcolor{yellow!40}24.29 & \cellcolor{orange!40}0.860 & 0.190 & 29.67 & 0.890 & 0.320\\
    \midrule 
    Ours & \cellcolor{red!40}29.90 & \cellcolor{orange!40}0.893 & \cellcolor{red!40}0.145 & \cellcolor{red!40}24.87 & \cellcolor{red!40} 0.873 & \cellcolor{red!40} 0.138 & \cellcolor{orange!40}30.07 & \cellcolor{red!40}0.907 & 0.247
\end{tabular}
\caption{
    \textbf{Comparison.} The red, orange and yellow colors represent the top three results. Competing metrics are extracted from respective papers, and ours are reported as the average of three runs.
}
\label{tab:comparisons}
\end{table*}
We first compare \sssName with the baselines on all scenes in their default settings, shown in \cref{tab:comparisons}. \sssName achieves overall the best results on 6 of the 9 metrics, and the second best on 2 metrics. The only exception is the LPIPS in Deep Blending, where the difference between \sssName and the best is 7$\times$10$^{-3}$. Furthermore, when investigating individual scenes, \sssName achieves the largest leading margin on Train. It achieves 23.23/0.844/0.170, where the second best method 3DHGS achieves 22.95/0.827/0.197, in PSNR, SSIM, and LPIPS, which is a 1.22\%/2.05\%/13.7\% improvement. Detailed scores for each scene are in the SM. We show qualitative comparison in Figure~\ref{fig:full_visual}.

\subsection{Parameter Efficiency}
\sssName has stronger representation power than 3DGS and its variants. The varying tail-thickness of its components enables \sssName to fit the data with fewer components, \ie higher parameter efficiency. We show this via experiments under different component numbers.

Since the SfM initialization gives different components in different scenes and a method normally increases the component number during learning, we introduce a coefficient to describe the latter as a multiplicity of the former. Denoting the initial component number as $\delta$, we test $\delta$, 1.4$\delta$, 1.8$\delta$, 2.2$\delta$, and 2.6$\delta$ as the maximum component number for comparison. Note even with 2.6$\delta$, the component number is still much smaller than the experiments in \cref{tab:comparisons}. Specifically, the 2.6$\delta$ vs the original 3DGS component number are 468k/1.1m, 364k/2.6m, 208k/3.4m, 96k/2.5m, 140k/5.9m, 520k/1.3m, 416k/1.2m, 364k/5.2m, 624k/1.8m, 286k/1.5m, 83k/4.75m, in Train, Truck, DrJohnson, Playroom, Bicycle, Bonsai, Counter, Garden, Kitchen, Room, Stump, corresponding to merely 42.5\%, 14\%, 6.1\%, 3.8\%, 2.4\%, 40\%, 34.7\%, 7\%, 34.7\%, 19.1\%, 1.7\% of the original components, a maximum of 98.3\% reduction.
\begin{figure*}
    \centering
    \def\svgwidth{1.0\linewidth}
    \fontsize{8pt}{10pt}\selectfont
\begingroup%
  \makeatletter%
  \providecommand\color[2][]{%
    \errmessage{(Inkscape) Color is used for the text in Inkscape, but the package 'color.sty' is not loaded}%
    \renewcommand\color[2][]{}%
  }%
  \providecommand\transparent[1]{%
    \errmessage{(Inkscape) Transparency is used (non-zero) for the text in Inkscape, but the package 'transparent.sty' is not loaded}%
    \renewcommand\transparent[1]{}%
  }%
  \providecommand\rotatebox[2]{#2}%
  \newcommand*\fsize{\dimexpr\f@size pt\relax}%
  \newcommand*\lineheight[1]{\fontsize{\fsize}{#1\fsize}\selectfont}%
  \ifx\svgwidth\undefined%
    \setlength{\unitlength}{1981.247078bp}%
    \ifx\svgscale\undefined%
      \relax%
    \else%
      \setlength{\unitlength}{\unitlength * \real{\svgscale}}%
    \fi%
  \else%
    \setlength{\unitlength}{\svgwidth}%
  \fi%
  \global\let\svgwidth\undefined%
  \global\let\svgscale\undefined%
  \makeatother%
  \begin{picture}(1,0.23466366)%
    \lineheight{1}%
    \setlength\tabcolsep{0pt}%
    \put(0.15702329,0.22311868){\makebox(0,0)[t]{\lineheight{1.25}\smash{\begin{tabular}[t]{c}(a) Mip-NeRF 360 Dataset\end{tabular}}}}%
    \put(0.49948616,0.22311868){\makebox(0,0)[t]{\lineheight{1.25}\smash{\begin{tabular}[t]{c}(b) Tanks \& Temples Dataset\end{tabular}}}}%
    \put(0.84194903,0.22315935){\makebox(0,0)[t]{\lineheight{1.25}\smash{\begin{tabular}[t]{c}(c) Deep Blending Dataset\end{tabular}}}}%
    \put(0,0){\includegraphics[width=\unitlength,page=1]{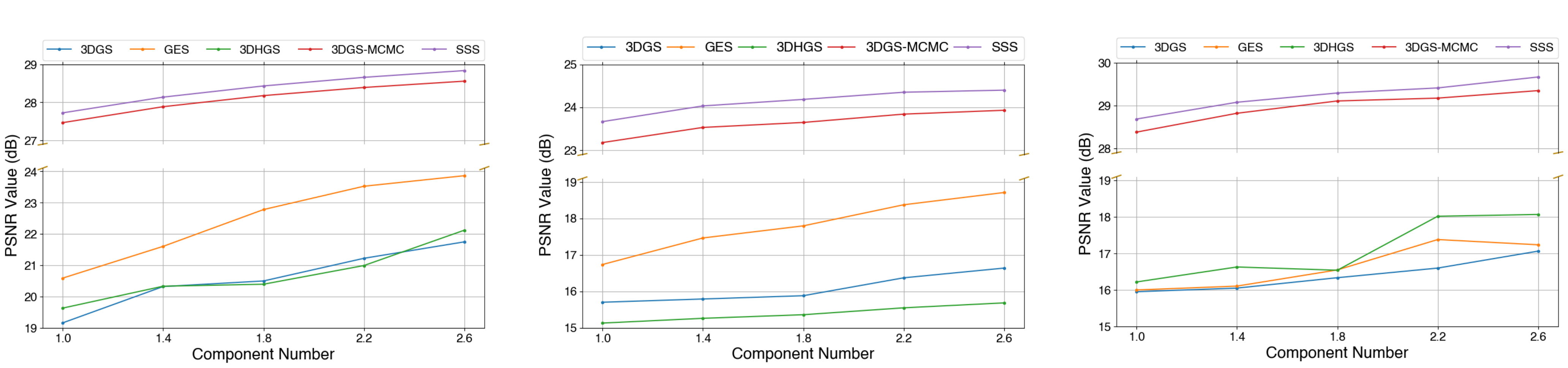}}%
  \end{picture}%
\endgroup%

    \caption{All methods with reduced component numbers.}
    \label{fig:para_effi}
\end{figure*}
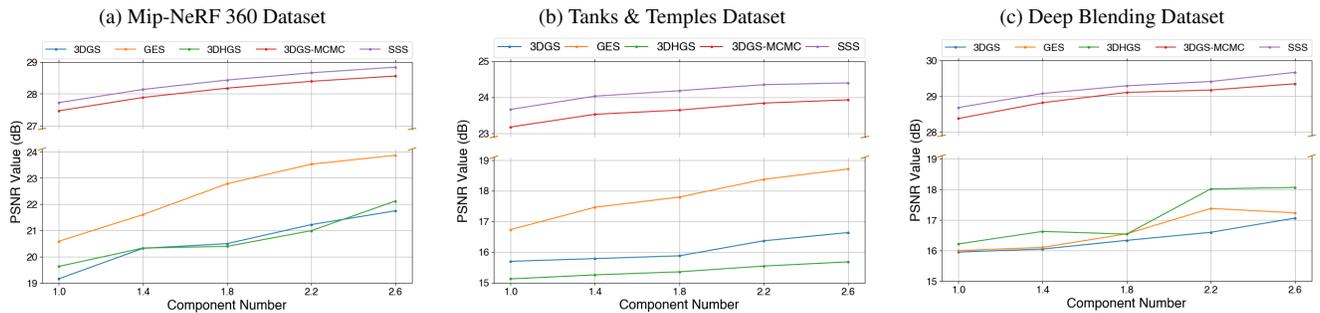

PSNR is averaged over the scenes in each dataset and shown in Figure~\ref{fig:para_effi}. First, \sssName outperforms all other methods in 15 out of 15 settings, demonstrating strong expressivity across all scenarios. Furthermore, when the component number decreases, all methods deteriorate, but \sssName deteriorates slowly comparatively, demonstrating that \sssName can fit the data much more efficiently than the rest. One specific example is the Tanks \& Temples in \cref{tab:comparisons}. \sssName achieves 23.6 PSNR with merely 180k components, already surpassing Mip-NeFR, 3DGS and GES. With around 300k components, \sssName achieves 24.4 PSNR, which is only slightly worse than 3DHGS and 3DGS-MCMC, by a margin at the scale of 10$^{-2}$. Note this is a comparison with the methods in \cref{tab:comparisons} where they use at least 1m components, \eg 3DGS employs around 1.1m and 2.6m in Train and Truck, while \sssName employs only around 364k and 468k, a maximum reduction of 82\% of the components.

\subsection{Qualitative Comparison}
\begin{figure*}[tb]
    \centering
    \def\svgwidth{1.0\linewidth}
    \fontsize{7pt}{9pt}\selectfont
    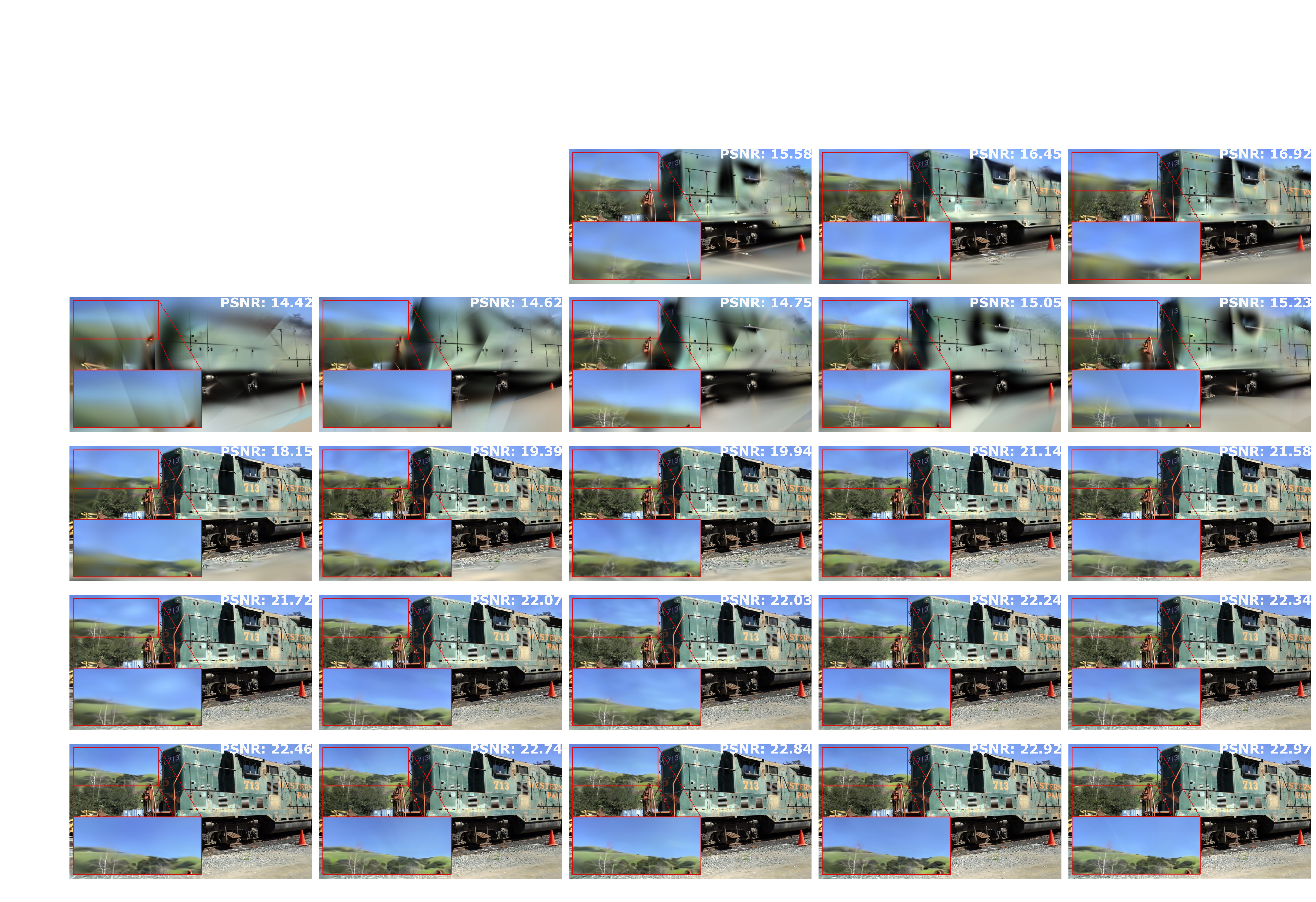
    \caption{\textbf{Visual results of all methods with varying component numbers.} In addition to the well-reconstructed main body of the train compared to other baselines, our method can use a small number of components to restore more details, such as plants on distant mountains, rocks on the ground nearby, etc. Besides, our sky has fewer noises and appears more similarly to the ground truth. Zoom in for details. Note that ours with 252k components has already achieved SOTA quality and beats most baselines.}
    \label{fig:para_effi_visual}
\end{figure*}
We further show a qualitative comparison in one scene across the different component numbers in \cref{fig:para_effi_visual}. The ground-truth is one view from the Train. When restricting the component number to around 180k, the original 3DGS and one of the state-of-the-art methods 3DHGS show significant blur. This is likely to be caused by the struggle between stretching Gaussians to cover large areas and narrowing them to reconstruct details, given a limited number of components at proposal. Due to the optimization relying on stochastic gradient descent, a local minimum is sought where the distribution of the Gaussians is sub-optimal. Intuitively, the issue can be mitigated by more flexible components and/or better optimization. As expected, this is shown in GES and 3DGS-MCMC, where the former employs a more flexible component (generalized exponential function) while the latter improves the optimization itself by MCMC. The improvements by both methods suggest that these are the correct directions to improve the paradigm of 3DGS.

Next, \sssName outperforms GES and 3DGS-MCMC visually when using 180k components. One example is the sky and the hill in the background in the left half of the image. GES creates a blurry background mixing the sky and the hill, with no discernible details, suggesting it uses a few components that are stretched to cover large areas. In contrast, 3DGS-MCMC can separate the sky from the hill. But it creates random white patches in the sky, which do not exist in the ground truth. This suggests that 3DGS-MCMC employs a relatively larger number of slim Gaussians to fit the details but meanwhile introduces additional noises. \sssName not only successfully separates the sky and the hill, but simultaneously retains the homogeneous color in the sky and reconstructs the details on the hill, \eg the trees and lawns. This is attributed to \sssName' capability of learning components with varying tail-fatness to adaptively capture large homogeneous areas and small heterogeneous regions.

Furthermore, when increasing the component number to 468k, all results are improved, as expected. 3DGS and 3DHGS still cannot work as well as other methods, suggesting they need more components. In addition, the difference between GES, 3DGS-MCMC, and \sssName starts to narrow. GES can separate the sky and the hill. Both GES and 3DGS-MCMC have fewer artifacts. However, as the component number increases, there are still noticeable noises introduced to the sky, which suggests that it is a common issue for both methods when using more components to cover an area with mixed homogeneous and heterogeneous regions. Comparatively, \sssName gives consistent performance across all component numbers, \ie clearly separating and reconstructing the homogeneous sky and the heterogeneous hill. Also, the visual quality from 180k to 468k does not change significantly for \sssName, but is noticeably improved for GES and 3DGS-MCMC, suggesting a higher parameter efficiency of \sssName in perception.

\paragraph{Ablation Study}
We conduct an ablation study to show the effectiveness of various components in SSS. These results prove how each component contributes to the final performance improvement. We give details in the SM.

\section{Conclusion, Discussion, and Future Work}
\label{sec:conclusion}
We proposed \sssNameLong (\sssName), a new non-monotonic mixture model, consisting of positive and negative Student's t distributions, learned by a principled SGHMC sampling. \sssName contains a simple yet strong and non-trivial generalization of 3DGS and its variants. \sssName outperforms existing methods in rendering quality, and shows high parameter efficiency, \eg achieving comparable quality with less than $\frac{1}{3}$ of components.

\sssName has limitations. Its primitives are restricted to symmetric and smooth t-distributions, limiting its representation. The sampling also needs hyperparameter tuning such as the percentage of negative components. In the future, we will combine other distribution families (\eg Laplace) with t-distribution to further enhance the expressivity, and make the SGHMC self-adaptive to achieve better balances between the positive and negative components.

\section*{Acknowledgement}
This project has received funding from the European Union’s Horizon 2020 research and innovation programme under grant agreement No 899739 CrowdDNA.

{
    \small
    \bibliographystyle{ieeenat_fullname}
    \bibliography{egbib}
}

\clearpage
\setcounter{section}{0}    
\setcounter{figure}{0}     
\setcounter{table}{0}      
\setcounter{equation}{0}   
\setcounter{page}{1}       
\twocolumn[
\begin{@twocolumnfalse}
    \centering 
    \Large\textbf{3D Student Splatting and Scooping\\ \vspace{0.1cm} Supplementary Material} \\ \vspace{0.25cm}
    \normalsize 
    \parbox{\textwidth}{
    \centering
    Jialin Zhu$^1$, Jiangbei Yue$^2$, Feixiang He$^1$, He Wang$^{1,3}$\\
    $^1$University College London, UK\ \ \ \ $^2$University of Leeds, UK\\
    $^3$AI Centre, University College London, UK
    } \vspace{0.2cm}
\end{@twocolumnfalse}
]

\section{Additional Experimental Results}
\begin{figure*}[tb]
    \centering
    \def\svgwidth{1.0\linewidth}
    \fontsize{8pt}{10pt}\selectfont
    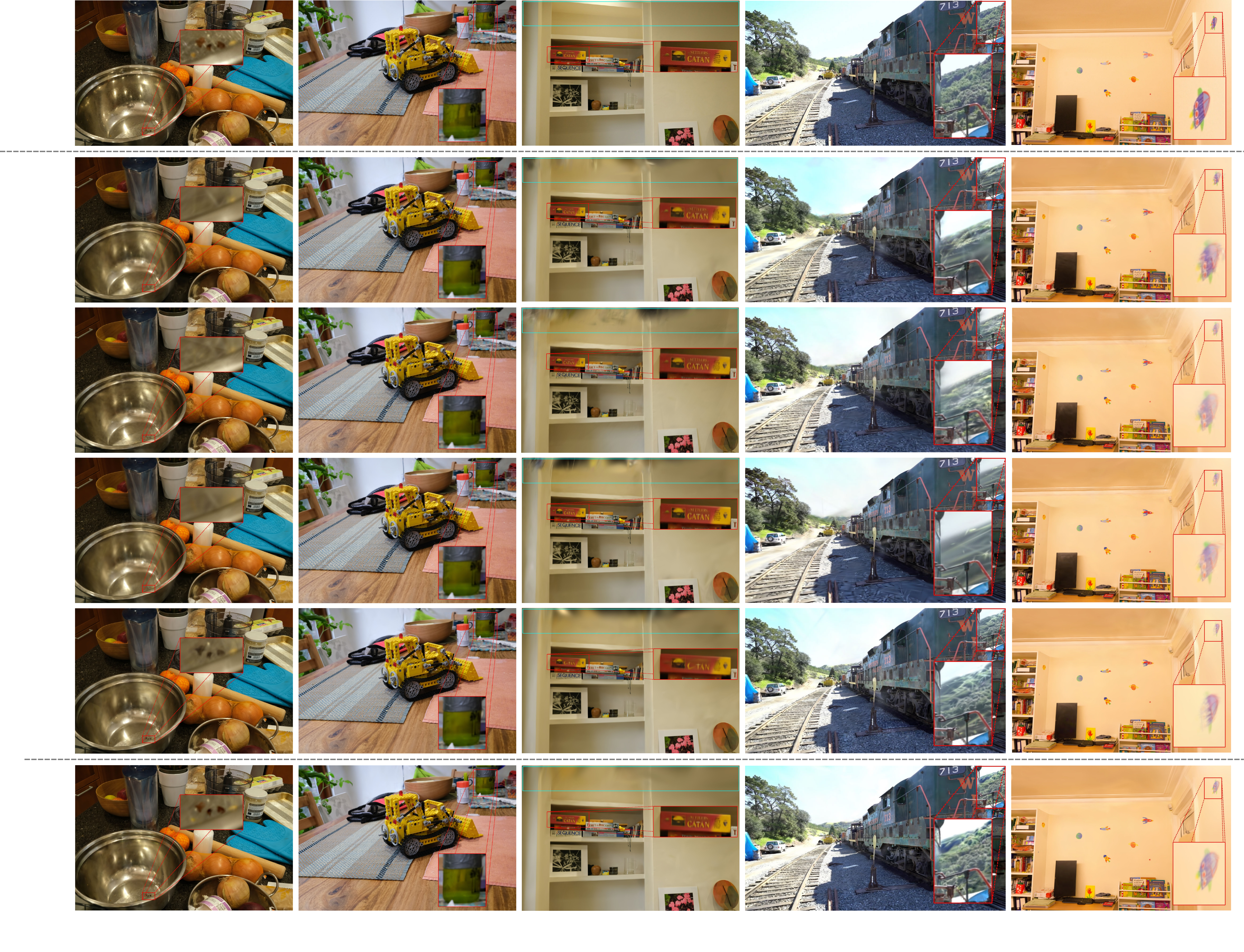
    \caption{\textbf{Visual comparison.} (a) SSS restores the best details inside the metal bowl. (b) SSS is the only one that can reconstruct the details of the chair refracted in the transparent cup. (c) The reconstruction of the wall edge (bright blue box) and the font were both done best by SSS. (d) SSS's details on the distant woods and the reconstruction of the sky are the best. (e) The reconstruction of the pattern on the wall is SSS at its best.}
    \label{fig:full_visual2}
\end{figure*}

We show more visual results of general benchmarks for Student Splatting and Scooping (SSS) against selected baselines: 3D Gaussian Splatting (3DGS)~\cite{kerbl20233d}, 3D Half-Gaussian Splatting (3DHGS)~\cite{li20243d}, Generalized Exponential Splattin (GES)~\cite{hamdi2024ges} and 3D Gaussian Splatting as Markov Chain Monte Carlo (3DGS-MCMC)~\cite{kheradmand20243d} in~\cref{fig:full_visual2}. For (a) in~\cref{fig:full_visual2}, there are a few food residues in the metal bowl. Only 3DGS-MCMC and SSS can successfully restore these details. 
Furthermore, SSS also achieves a better reconstruction of colors than 3DGS-MCMC with these residues. SSS is the only method that can reproduce the sharp details of the chair refracted in the transparent water glass in~\cref{fig:full_visual2} (b). For (c) in ~\cref{fig:full_visual2}, the difficulty lies in reconstructing the details of the upper wall edge and the items on the cabinet (texts, etc.). Considering these two difficulties, SSS performs best in this scenario. SSS can also ensure that the details at the edge of the image are restored to the greatest extent, which is reflected in (d) and (e) of ~\cref{fig:full_visual2}. In addition, SSS is also the best for reconstructing pure color areas (sky, wall, etc.).

\begin{figure*}[tb]
    \centering
    \def\svgwidth{1.0\linewidth}
    \fontsize{8pt}{10pt}\selectfont
    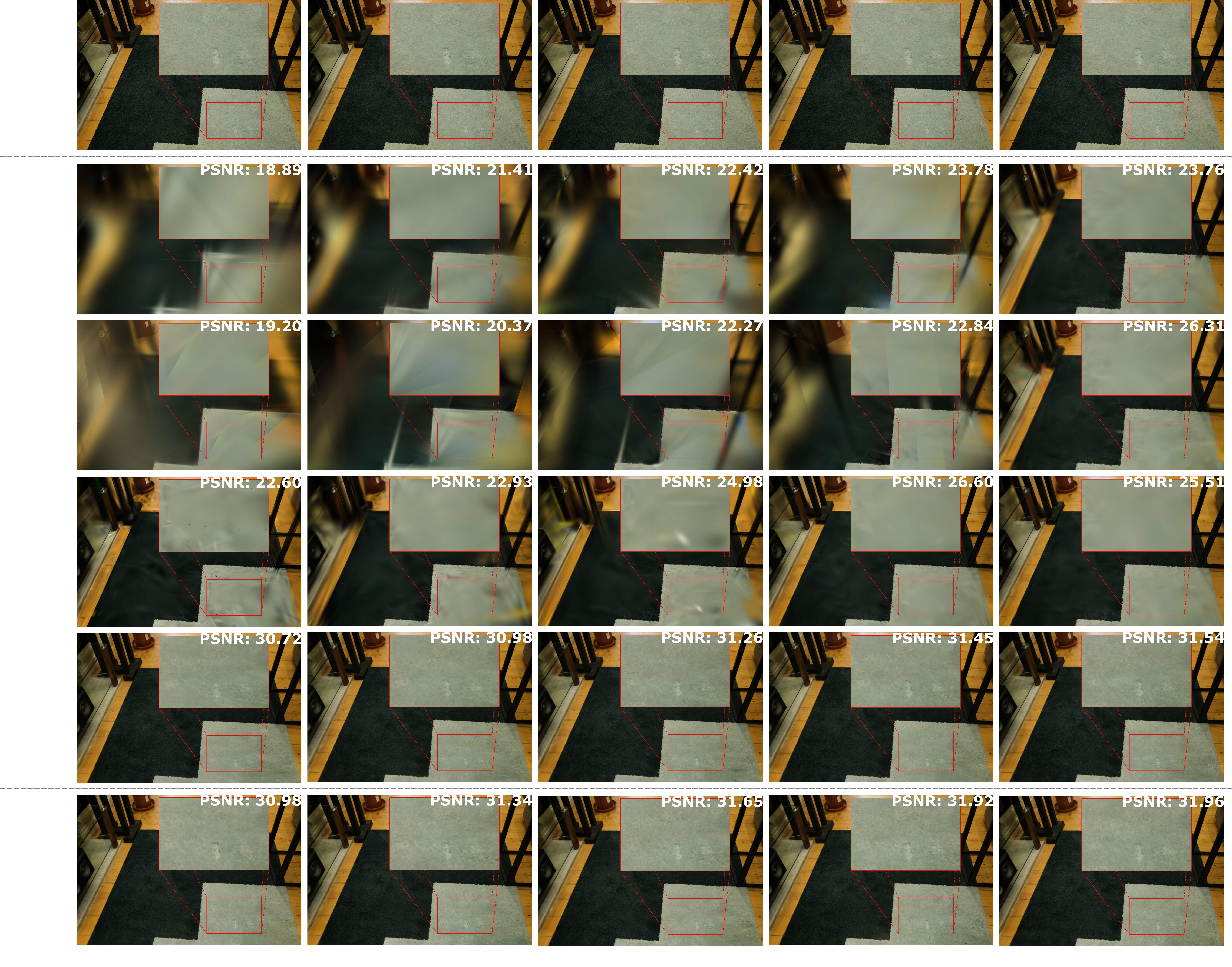
    \caption{\textbf{Visual results of all methods with varying component numbers of room scene from Mip-NeRF 360.} Only 3DGS-MCMC and SSS can restore the details of the carpet, but the result of SSS obviously has a more realistic carpet texture.}
    \label{fig:para_effi_visual2}
\end{figure*}

\begin{figure*}[tb]
    \centering
    \def\svgwidth{1.0\linewidth}
    \fontsize{8pt}{10pt}\selectfont
    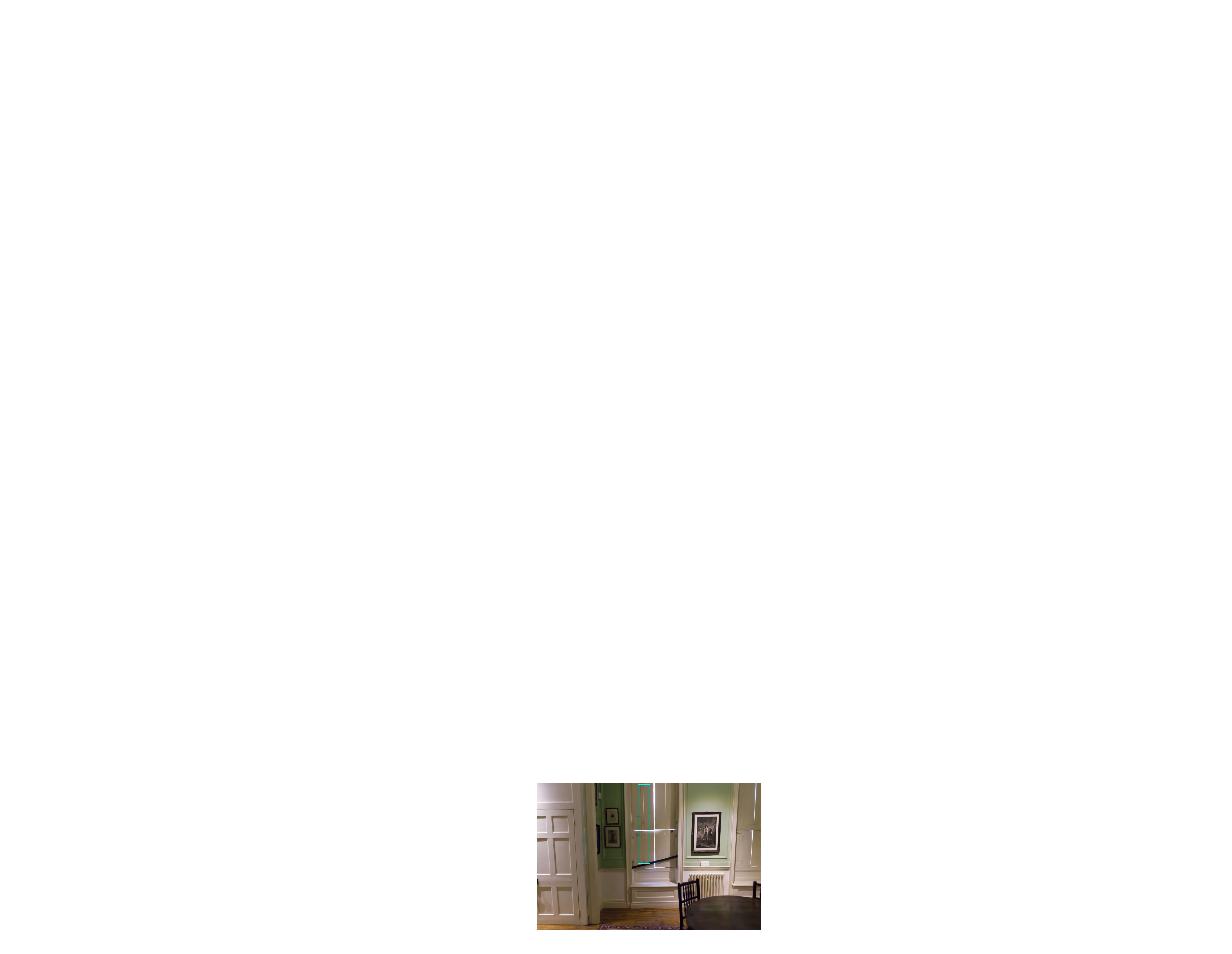
    \caption{\textbf{Visual results of all methods with varying component numbers of drjohnson scene from Deep Blending.} 3DGS, 3DHSGS, and GES are completely unable to reconstruct quality results. The results of 3DGS-MCMC and SSS are relatively better. SSS can restore more details (such as the window gaps in the blue box) than 3DGS-MCMC.}
    \label{fig:para_effi_visual3}
\end{figure*}

We have illustrated the results of the Train scene in the Tanks \& Temples dataset~\cite{knapitsch2017tanks} in the main context for the varying component numbers experiment. Here we further show two more scenes (room from Mip-MeRF 360~\cite{barron2022mip} and drjohnson from Deep Blending~\cite{hedman2018deep}) in~\cref{fig:para_effi_visual2,fig:para_effi_visual3}.
In~\cref{fig:para_effi_visual2}, the difficulty of 3D reconstruction lies in the texture of the carpet. Both 3DGS, 3DHGS, and GES can only restore the approximate shape but cannot take into account the details when the number of components is small. 3DGS-MCMC and SSS can restore the details of the carpet. However, because SSS has a scooping operation, the carpet it reconstructs has a more realistic texture. 3DGS, 3DHGS, and GES are completely unable to reconstruct quality results in the setting of 80k to 208k components in~\cref{fig:para_effi_visual3}. The reconstruction of 3DGS-MCMC is good enough but loses some details (window gaps inside the basket marked area), while SSS is still the best method for capturing details.

\subsection{Detailed Results on Each Scene}
We show detailed comparisons between our method and the baselines on every scene on Peak Signal-to-Noise Ratio (PSNR), Structural Similarity Index Metric (SSIM)~\cite{wang2004image}, and Learned Perceptual Image Patch Similarity (LPIPS)~\cite{zhang2018unreasonable} metrics among three datasets (Mip-NeRF 360~\cite{barron2022mip}, Tanks \& Temples~\cite{knapitsch2017tanks}, Deep Blending~\cite{hedman2018deep}) in~\cref{tab:comparisons_360_psnr,tab:comparisons_360_ssim,tab:comparisons_360_lpips,tab:comparisons_td_psnr,tab:comparisons_td_ssim,tab:comparisons_td_lpips}. 

Note we only include baselines that provide detailed evaluation scores on each scene. These baselines are Mip-NeRF~\cite{barron2021mip}, 3DGS~\cite{kerbl20233d}, Scaffold-GS~\cite{lu2024scaffold}, 3DHGS~\cite{li20243d} and 3DGS-MCMC~\cite{kheradmand20243d}. This is due to the intrinsic randomness in the training of these methods. When we re-train the models ourselves and often obtain slightly different results from their papers. Therefore, we use the results reported in their original papers.

Overall, our method outperforms all the baselines. A close second is 3DGS-MCMC, which is a state-of-the-art model recently. There are some baselines that outperform both our method and 3DGS-MCMC on individual scenes under some metrics, but overall, our method and 3DGS-MCMC are the best and second best methods.

Beyond the selected methods for baseline, we do realize that there might be other methods that are not included here but achieve higher scores on certain scenes and metrics, especially when it comes to specific application settings, \eg 3D reconstruction. However, our goal here is to restrict our comparison to the methods that aim to improve 3DGS on its fundamental formulation and can be potentially used as a generic-purposed component.

\begin{table}
\resizebox{0.48\textwidth}{!}{
\begin{tabular}{l|ccccccc|c}
    Method $\backslash$ Scene & bicycle & bonsai & counter & garden & kitchen & room & stump & average\\
    \hline 
    3DGS & 25.25 & 31.98 & 28.70 & 27.41 & 30.32 & 30.63 & 26.55 & 28.69 \\
    Mip-NeRF & 24.37 & \cellcolor{orange!40}33.46 & \cellcolor{yellow!40}29.55 & 26.98 & \cellcolor{yellow!40}32.23 & 31.64 & 26.40 & 29.23 \\
    Scaffold-GS & 24.50 & 32.70 & 29.34 & 27.17 & 31.30 & 31.93 & 26.27 & 28.84\\
    3DHGS & \cellcolor{yellow!40}25.39 & \cellcolor{yellow!40}33.30 & \cellcolor{orange!40}29.62 & \cellcolor{yellow!40}27.68 & 32.17 & \cellcolor{yellow!40}32.12 & \cellcolor{yellow!40}26.64 & \cellcolor{yellow!40}29.56\\
    3DGS-MCMC & \cellcolor{red!40}26.15 & 32.88 & 29.51 & \cellcolor{red!40}28.16 & \cellcolor{orange!40}32.27 & \cellcolor{orange!40}32.48 & \cellcolor{red!40}27.80 & \cellcolor{orange!40}29.89\\
    \hline
    SSS & \cellcolor{orange!40}25.68 & \cellcolor{red!40}33.50 & \cellcolor{red!40}29.87 & \cellcolor{orange!40}28.09 & \cellcolor{red!40}32.43& \cellcolor{red!40}32.57 & \cellcolor{orange!40}27.17 & \cellcolor{red!40}29.90\\
\end{tabular}
}
\caption{
    \textbf{PSNR results for every scene in Mip-NeRF 360 dataset.} The red, orange, and yellow colors represent the top three results. Competing metrics are extracted from respective papers, and ours are reported as the average of three runs.
}
\label{tab:comparisons_360_psnr}
\end{table}

\begin{table}
\resizebox{0.48\textwidth}{!}{
\begin{tabular}{l|ccccccc|c}
    Method $\backslash$ Scene & bicycle & bonsai & counter & garden & kitchen & room & stump & average\\
    \hline 
    3DGS & \cellcolor{yellow!40}0.771 & 0.938 & 0.905 & \cellcolor{yellow!40}0.868 & 0.922 & 0.914 & 0.775 & 0.870\\
    Mip-NeRF & 0.685 & 0.941 & 0.894 & 0.813 & 0.920 & 0.913 & 0.744 & 0.844\\
    Scaffold-GS & 0.705 & \cellcolor{yellow!40}0.946 & \cellcolor{yellow!40}0.914 & 0.842 & 0.928 & \cellcolor{yellow!40}0.925 & \cellcolor{yellow!40}0.784 & 0.848\\
    3DHGS & 0.768 & \cellcolor{orange!40}0.950 & 0.909 & \cellcolor{yellow!40}0.868 & \cellcolor{yellow!40}0.930 & 0.921 & 0.770 & \cellcolor{yellow!40}0.873\\
    3DGS-MCMC & \cellcolor{red!40}0.810 & \cellcolor{orange!40}0.950 & \cellcolor{orange!40}0.920 & \cellcolor{red!40}0.890 & \cellcolor{red!40}0.940 & \cellcolor{red!40}0.940 & \cellcolor{red!40}0.820 & \cellcolor{red!40}0.900\\
    \hline
    SSS & \cellcolor{orange!40}0.798 & \cellcolor{red!40}0.956 & \cellcolor{red!40}0.926 & \cellcolor{orange!40}0.882 & \cellcolor{orange!40}0.939 & \cellcolor{orange!40}0.938 & \cellcolor{orange!40}0.813 & \cellcolor{orange!40}0.893\\
\end{tabular}
}
\caption{
    \textbf{SSIM results for every scene in Mip-NeRF 360 dataset.} The red, orange, and yellow colors represent the top three results. Competing metrics are extracted from respective papers, and ours are reported as the average of three runs.
}
\label{tab:comparisons_360_ssim}
\end{table}

\begin{table}
\resizebox{0.48\textwidth}{!}{
\begin{tabular}{l|ccccccc|c}
    Method $\backslash$ Scene & bicycle & bonsai & counter & garden & kitchen & room & stump & average\\
    \hline 
    3DGS & 0.205 & 0.205 & 0.204 & \cellcolor{yellow!40}0.103 & 0.129 & 0.220 & \cellcolor{yellow!40}0.210 & \cellcolor{yellow!40}0.182\\
    Mip-NeRF & 0.301 & \cellcolor{orange!40}0.176 & 0.204 & 0.170 & 0.127 & \cellcolor{yellow!40}0.211 & 0.261 & 0.207\\
    Scaffold-GS & 0.306 & 0.185 & \cellcolor{orange!40}0.191 & 0.146 & \cellcolor{yellow!40}0.126 & \cellcolor{orange!40}0.202 & 0.284 & 0.220\\
    3DHGS & \cellcolor{yellow!40}0.202 & \cellcolor{yellow!40}0.180 & \cellcolor{yellow!40}0.201 & 0.104 & \cellcolor{orange!40}0.125 & 0.220 & 0.215 & \cellcolor{orange!40}0.178\\
    3DGS-MCMC & \cellcolor{orange!40}0.180 & 0.220 & 0.220 & \cellcolor{orange!40}0.100 & 0.140 & 0.250 & \cellcolor{orange!40}0.190 & 0.190\\
    \hline
    SSS & \cellcolor{red!40}0.173 & \cellcolor{red!40}0.151 & \cellcolor{red!40}0.156 & \cellcolor{red!40}0.009 & \cellcolor{red!40}0.104 & \cellcolor{red!40}0.167 & \cellcolor{red!40}0.174 & \cellcolor{red!40}0.145\\
\end{tabular}
}
\caption{
    \textbf{LPIPS results for every scene in Mip-NeRF 360 dataset.} The red, orange, and yellow colors represent the top three results. Competing metrics are extracted from respective papers, and ours are reported as the average of three runs.
}
\label{tab:comparisons_360_lpips}
\end{table}

\begin{table}
\resizebox{0.48\textwidth}{!}{
\begin{tabular}{l|ccccccc}
    Dataset - Scene & \multicolumn{3}{c|}{Tanks\&Temples} & \multicolumn{3}{c}{Deep Blending}\\
    Method & train & truck & average & drjohnson & playroom & average\\
    \hline 
    3DGS & 21.09 & 25.18 & 23.14  & 28.77 & 30.04 & 29.41 \\
    Mip-NeRF & 19.52 & 24.91 & 22.22 & 29.14 & 29.66 & 29.40 \\
    Scaffold-GS & 22.15 & 25.77 & 23.96 & \cellcolor{red!40}29.80 & \cellcolor{red!40}30.62 & \cellcolor{red!40}30.21 \\
    3DHGS & \cellcolor{orange!40}22.95 & \cellcolor{yellow!40}26.04 & \cellcolor{orange!40}24.49  &\cellcolor{yellow!40}29.32 & 30.20 & \cellcolor{yellow!40}29.76 \\
    3DGS-MCMC & \cellcolor{yellow!40}22.47 & \cellcolor{orange!40}26.11 & \cellcolor{yellow!40}24.29 & 29.00 & \cellcolor{yellow!40}30.33 & 29.67 \\
    \hline
    SSS & \cellcolor{red!40}23.32 & \cellcolor{red!40}26.41 & \cellcolor{red!40}24.87 & \cellcolor{orange!40}29.66 & \cellcolor{orange!40}30.47 & \cellcolor{orange!40}30.07 \\
\end{tabular}
}
\caption{
    \textbf{PSNR results for every scene in Tanks \& Temples and Deep Blending dataset.} The red, orange, and yellow colors represent the top three results. Competing metrics are extracted from respective papers, and ours are reported as the average of three runs.
}
\label{tab:comparisons_td_psnr}
\end{table}

\begin{table}
\resizebox{0.48\textwidth}{!}{
\begin{tabular}{l|ccccccc}
    Dataset - Scene & \multicolumn{3}{c|}{Tanks\&Temples} & \multicolumn{3}{c}{Deep Blending}\\
    Method & train & truck & average & drjohnson & playroom & average\\
    \hline 
    3DGS & 0.802 & 0.879 & 0.841 & 0.899 & \cellcolor{yellow!40}0.906 & 0.903\\
    Mip-NeRF & 0.660 & 0.857 & 0.759 & 0.901 & 0.900 & 0.901 \\
    Scaffold-GS & 0.822 & 0.883 & 0.853 & \cellcolor{red!40}0.907 & 0.904 & \cellcolor{orange!40}0.906 \\
    3DHGS & \cellcolor{yellow!40}0.827 & \cellcolor{yellow!40}0.887 & \cellcolor{yellow!40}0.857 & \cellcolor{yellow!40}0.904 & \cellcolor{orange!40}0.907 & \cellcolor{yellow!40}0.905 \\
    3DGS-MCMC & \cellcolor{orange!40}0.830 & \cellcolor{orange!40}0.890 & \cellcolor{orange!40}0.860 & 0.890 & 0.900 & 0.890\\
    \hline
    SSS & \cellcolor{red!40}0.850 & \cellcolor{red!40}0.897 & \cellcolor{red!40}0.873 & \cellcolor{orange!40}0.905 & \cellcolor{red!40}0.909 & \cellcolor{red!40}0.907 \\
\end{tabular}
}
\caption{
    \textbf{SSIM results for every scene in Tanks \& Temples and Deep Blending dataset.} The red, orange, and yellow colors represent the top three results. Competing metrics are extracted from respective papers, and ours are reported as the average of three runs.
}
\label{tab:comparisons_td_ssim}
\end{table}

\begin{table}
\resizebox{0.48\textwidth}{!}{
\begin{tabular}{l|ccccccc}
    Dataset - Scene & \multicolumn{3}{c|}{Tanks\&Temples} & \multicolumn{3}{c}{Deep Blending}\\
    Method & train & truck & drjohnson & playroom\\
    \hline 
    3DGS & 0.218 & 0.148 & 0.183 & \cellcolor{yellow!40}0.244 & \cellcolor{red!40}0.241 & \cellcolor{orange!40}0.243\\
    Mip-NeRF & 0.354 & 0.159 & 0.257 & \cellcolor{red!40}0.237 & 0.252 & \cellcolor{yellow!40}0.245\\
    Scaffold-GS & \cellcolor{yellow!40}0.206 & 0.147 & \cellcolor{yellow!40}0.177 & 0.250 & 0.258 & 0.254\\
    3DHGS & \cellcolor{orange!40}0.197 & \cellcolor{yellow!40}0.141 & \cellcolor{orange!40}0.169 & \cellcolor{orange!40}0.240 & \cellcolor{orange!40}0.243 & \cellcolor{red!40}0.242\\
    3DGS-MCMC & 0.240 & \cellcolor{orange!40}0.140 & 0.860 & 0.330 & 0.310 & 0.320\\
    \hline
    SSS & \cellcolor{red!40}0.166 & \cellcolor{red!40}0.109 & \cellcolor{red!40}0.138 & 0.249 & \cellcolor{yellow!40}0.245 & 0.247\\
\end{tabular}
}
\caption{
    \textbf{LPIPS results for every scene in Tanks \& Temples and Deep Blending dataset.} The red, orange, and yellow colors represent the top three results. Competing metrics are extracted from respective papers, and ours are reported as the average of three runs.
}
\label{tab:comparisons_td_lpips}
\end{table}

\subsection{Detailed Results on Varying Component Numbers}
We show detailed results of varying component numbers, on every scene with PSNR, SSIM, and LPIPS in Mip-NeRF 360, Tanks \& Temples, and Deep Blending datasets. Since we need to re-train all methods for comparison, we ensure the implementation is as fair as possible. Based on whether codes are open sourced and furthermore how easily they can be adapted for comparison (explained later), we selected 3DGS~\cite{kerbl20233d}, GES~\cite{hamdi2024ges}, 3DHGS~\cite{li20243d} and 3DGS-MCMC~\cite{kheradmand20243d} as the baselines. We use $\delta$ to represent the number of the initial components from Structure from Motion (SFM) reconstruction~\cite{ullman1979interpretation} for different scenes.

The results are shown in~\cref{tab:varying_360_psnr,tab:varying_360_ssim,tab:varying_360_lpips,tab:varying_td_psnr,tab:varying_td_ssim,tab:varying_td_lpips}. In total, there are 11 (scenes) $\times$ 5 (component numbers) $\times$ 3 (metrics) = 165 comparisons. For the absolute majority, SSS achieves the best. Furthermore, when it is not the best method, it is the close second to 3DGS-MCMC. This is an exhaustive comparison of many datasets, metrics, and more importantly different numbers of components.

\begin{table}
\resizebox{0.47\textwidth}{!}{
\begin{tabular}{l|ccccccc}
    Scene - Components & \multicolumn{5}{c|}{Mip-NeRF 360 - average}\\
    Method & $\delta$ & $1.4\delta$ & $1.8\delta$ & $2.2\delta$ & $2.6\delta$\\
    \hline
    3DGS & 19.16 & 20.32 & 20.50 & 21.22 & 21.75\\
    GES & \cellcolor{yellow!40}20.59 & \cellcolor{yellow!40}21.60 & \cellcolor{yellow!40}22.78 & \cellcolor{yellow!40}23.52 & \cellcolor{yellow!40}23.86\\
    3DHGS & 19.63 & 20.33 & 20.39 & 20.99 & 22.12\\
    3DGS-MCMC & \cellcolor{orange!40}27.47 & \cellcolor{orange!40}27.89 & \cellcolor{orange!40}28.18 & \cellcolor{orange!40}28.39 & \cellcolor{orange!40}28.56\\
    SSS & \cellcolor{red!40}27.72 & \cellcolor{red!40}28.14 & \cellcolor{red!40}28.43 & \cellcolor{red!40}28.66 & \cellcolor{red!40}28.84\\
    \hline
    Scene - Components & \multicolumn{5}{c|}{Mip-NeRF 360 - bicycle}\\
    Method & 54k & 75k & 97k & 118k & 140k\\
    \hline 
    3DGS & 18.32 & 18.53 & \cellcolor{yellow!40}18.62 & \cellcolor{yellow!40}18.75 & \cellcolor{yellow!40}18.52\\
    GES & 18.31 & 18.29 & 18.20 & 18.13 & 18.19\\
    3DHGS & \cellcolor{yellow!40}18.45 & \cellcolor{yellow!40}18.72 & \cellcolor{yellow!40}18.62 & 18.61 & 18.50\\
    3DGS-MCMC & \cellcolor{red!40}23.04 & \cellcolor{red!40}23.42 & \cellcolor{red!40}23.74 & \cellcolor{red!40}23.97 & \cellcolor{red!40}24.17\\
    SSS & \cellcolor{orange!40}22.97 & \cellcolor{orange!40}23.29 & \cellcolor{orange!40}23.53 & \cellcolor{orange!40}23.71 & \cellcolor{orange!40}23.95\\
    \hline
    Scene - Components & \multicolumn{5}{c|}{Mip-NeRF 360 - bonsai}\\
    Method & 206k & 280k & 360k & 440k & 520k\\
    \hline
    3DGS & 20.50 & \cellcolor{yellow!40}22.93 & 23.39 & 23.97 & 24.54\\
    GES & \cellcolor{yellow!40}22.47 & 24.12 & \cellcolor{yellow!40}25.51 & \cellcolor{yellow!40}27.02 & \cellcolor{yellow!40}29.27\\
    3DHGS & 21.37 & 22.90 & 23.45 & 24.62 & 24.82\\
    3DGS-MCMC & \cellcolor{orange!40}31.14 & \cellcolor{orange!40}31.56 & \cellcolor{orange!40}31.85 & \cellcolor{orange!40}32.03 & \cellcolor{orange!40}32.17\\
    SSS & \cellcolor{red!40}31.67 & \cellcolor{red!40}32.21 & \cellcolor{red!40}32.52 & \cellcolor{red!40}32.75 & \cellcolor{red!40}32.94\\
    \hline
    Scene - Components & \multicolumn{5}{c|}{Mip-NeRF 360 - counter}\\
    Method & 155k & 224k & 288k & 352k & 416k\\
    \hline
    3DGS & 18.26 & 19.07 & 18.84 & 20.78 & 23.73\\
    GES & \cellcolor{yellow!40}19.68 & \cellcolor{yellow!40}23.64 & \cellcolor{yellow!40}26.35 & \cellcolor{yellow!40}26.74 & \cellcolor{yellow!40}27.45\\
    3DHGS & 17.77 & 18.36 & 18.89 & 20.48 & 23.59\\
    3DGS-MCMC & \cellcolor{orange!40}28.38 & \cellcolor{orange!40}28.68 & \cellcolor{orange!40}28.82 & \cellcolor{orange!40}28.93 & \cellcolor{orange!40}29.03\\
    SSS & \cellcolor{red!40}28.71 & \cellcolor{red!40}29.06 & \cellcolor{red!40}29.21 & \cellcolor{red!40}29.39 & \cellcolor{red!40}29.48\\
    \hline
    Scene - Components & \multicolumn{5}{c|}{Mip-NeRF 360 - garden}\\
    Method & 138k & 196k & 252k & 308k & 364k\\
    \hline
    3DGS & 16.73 & 17.34 & 17.49 & 17.81 & 18.05\\
    GES & 16.62 & 16.94 & 17.24 & 17.58 & 17.92\\
    3DHGS & \cellcolor{yellow!40}17.36 & \cellcolor{yellow!40}17.44 & \cellcolor{yellow!40}17.70 & \cellcolor{yellow!40}18.25 & \cellcolor{yellow!40}18.49\\
    3DGS-MCMC & \cellcolor{orange!40}25.23 & \cellcolor{orange!40}25.70 & \cellcolor{orange!40}26.02 & \cellcolor{orange!40}26.26 & \cellcolor{orange!40}26.47\\
    SSS & \cellcolor{red!40}25.32 & \cellcolor{red!40}25.80 & \cellcolor{red!40}26.17 & \cellcolor{red!40}26.41 & \cellcolor{red!40}26.64\\
    \hline
    Scene - Components & \multicolumn{5}{c|}{Mip-NeRF 360 - kitchen}\\
    Method & 241k & 336k & 432k & 528k & 624k\\
    \hline
    3DGS & 21.19 & 22.70 & 22.10 & 22.63 & 22.70\\
    GES & \cellcolor{yellow!40}24.78 & \cellcolor{yellow!40}25.74 & \cellcolor{yellow!40}27.65 & \cellcolor{yellow!40}29.34 & \cellcolor{yellow!40}29.54\\
    3DHGS & 23.19 & 24.20 & 21.40 & 21.61 & 22.60\\
    3DGS-MCMC & \cellcolor{orange!40}30.11 & \cellcolor{orange!40}30.78 & \cellcolor{orange!40}31.05 & \cellcolor{orange!40}31.26 & \cellcolor{orange!40}31.45\\
    SSS & \cellcolor{red!40}30.84 & \cellcolor{red!40}31.33 & \cellcolor{red!40}31.64 & \cellcolor{red!40}31.77 & \cellcolor{red!40}31.97\\
    \hline
    Scene - Components & \multicolumn{5}{c|}{Mip-NeRF 360 - room}\\
    Method & 112k & 154k & 198k & 242k & 286k\\
    \hline
    3DGS & 18.89 & 21.41 & 22.42 & 23.78 & 23.76\\
    GES & \cellcolor{yellow!40}22.60 & \cellcolor{yellow!40}22.93 & \cellcolor{yellow!40}24.98 & \cellcolor{yellow!40}26.60 & \cellcolor{yellow!40}25.51\\
    3DHGS & 19.20 & 20.37 & 22.27 & 22.83 & 26.31\\
    3DGS-MCMC & \cellcolor{orange!40}30.72 & \cellcolor{orange!40}30.98 & \cellcolor{orange!40}31.26 & \cellcolor{orange!40}31.45 & \cellcolor{orange!40}31.54\\
    SSS & \cellcolor{red!40}30.98 & \cellcolor{red!40}31.34 & \cellcolor{red!40}31.65 & \cellcolor{red!40}31.92 & \cellcolor{red!40}31.96\\
    \hline
    Scene - Components & \multicolumn{5}{c|}{Mip-NeRF 360 - stump}\\
    Method & 32k & 44.8k & 57.6k & 70.4k & 83.2k\\
    \hline
    3DGS & \cellcolor{yellow!40}20.21 & 20.26 & \cellcolor{yellow!40}20.61 & \cellcolor{yellow!40}20.82 & \cellcolor{yellow!40}20.92\\
    GES & 19.67 & 19.55 & 19.51 & 19.25 & 19.13\\
    3DHGS & 20.08 & \cellcolor{yellow!40}20.31 & 20.42 & 20.52 & 20.53\\
    3DGS-MCMC & \cellcolor{red!40}23.64 & \cellcolor{red!40}24.08 & \cellcolor{red!40}24.51 & \cellcolor{red!40}24.84 & \cellcolor{red!40}25.09\\
    SSS & \cellcolor{orange!40}23.57 & \cellcolor{orange!40}23.95 & \cellcolor{orange!40}24.31 & \cellcolor{orange!40}24.66 & \cellcolor{orange!40}24.91\\
\end{tabular}
}
\caption{
    \textbf{PSNR results of varying component numbers experiments for every scene in the Mip-NeRF 360 dataset.}
}
\label{tab:varying_360_psnr}
\end{table}

\begin{table}
\resizebox{0.47\textwidth}{!}{
\begin{tabular}{l|ccccccc}
    Scene - Components & \multicolumn{5}{c|}{Mip-NeRF 360 - average}\\
    Method & $\delta$ & $1.4\delta$ & $1.8\delta$ & $2.2\delta$ & $2.6\delta$\\
    \hline
    3DGS & 0.584 & 0.609 & 0.610 & 0.626 & 0.640\\
    GES & \cellcolor{yellow!40}0.612 & \cellcolor{yellow!40}0.630 & \cellcolor{yellow!40}0.652 & \cellcolor{yellow!40}0.667 & \cellcolor{yellow!40}0.674\\
    3DHGS & 0.591 & 0.603 & 0.599 & 0.611 & 0.634\\
    3DGS-MCMC & \cellcolor{red!40}0.803 & \cellcolor{red!40}0.818 & \cellcolor{red!40}0.829 & \cellcolor{red!40}0.837 & \cellcolor{red!40}0.844\\
    SSS & \cellcolor{orange!40}0.802 & \cellcolor{orange!40}0.816 & \cellcolor{orange!40}0.827 & \cellcolor{orange!40}0.835 & \cellcolor{orange!40}0.842\\
    \hline
    Scene - Components & \multicolumn{5}{c|}{Mip-NeRF 360 - bicycle}\\
    Method & 54k & 75k & 97k & 118k & 140k\\
    \hline 
    3DGS & 0.367 & \cellcolor{yellow!40}0.375 & \cellcolor{yellow!40}0.377 & \cellcolor{yellow!40}0.382 & \cellcolor{yellow!40}0.377\\
    GES & \cellcolor{yellow!40}0.372 & 0.369 & 0.368 & 0.368 & 0.370\\
    3DHGS & 0.369 & 0.374 & 0.372 & 0.376 & 0.373\\
    3DGS-MCMC & \cellcolor{red!40}0.603 & \cellcolor{red!40}0.630 & \cellcolor{red!40}0.650 & \cellcolor{red!40}0.666 & \cellcolor{red!40}0.679\\
    SSS & \cellcolor{orange!40}0.597 & \cellcolor{orange!40}0.620 & \cellcolor{orange!40}0.639 & \cellcolor{orange!40}0.652 & \cellcolor{orange!40}0.666\\
    \hline
    Scene - Components & \multicolumn{5}{c|}{Mip-NeRF 360 - bonsai}\\
    Method & 206k & 280k & 360k & 440k & 520k\\
    \hline
    3DGS & 0.790 & 0.822 & 0.820 & 0.828 & 0.838\\
    GES & \cellcolor{yellow!40}0.811 & \cellcolor{yellow!40}0.836 & \cellcolor{yellow!40}0.861 & \cellcolor{yellow!40}0.882 & \cellcolor{yellow!40}0.911\\
    3DHGS & 0.802 & 0.821 & 0.827 & 0.837 & 0.839\\
    3DGS-MCMC & \cellcolor{red!40}0.938 & \cellcolor{orange!40}0.942 & \cellcolor{orange!40}0.945 & \cellcolor{orange!40}0.947 & \cellcolor{orange!40}0.948\\
    SSS & \cellcolor{red!40}0.938 & \cellcolor{red!40}0.944 & \cellcolor{red!40}0.948 & \cellcolor{red!40}0.950 & \cellcolor{red!40}0.951\\
    \hline
    Scene - Components & \multicolumn{5}{c|}{Mip-NeRF 360 - counter}\\
    Method & 155k & 224k & 288k & 352k & 416k\\
    \hline
    3DGS & 0.663 & 0.686 & 0.682 & 0.731 & 0.799\\
    GES & \cellcolor{yellow!40}0.696 & \cellcolor{yellow!40}0.785 & \cellcolor{yellow!40}0.854 & \cellcolor{yellow!40}0.866 & \cellcolor{yellow!40}0.882\\
    3DHGS & 0.650 & 0.667 & 0.680 & 0.719 & 0.788\\
    3DGS-MCMC & \cellcolor{red!40}0.901 & \cellcolor{red!40}0.907 & \cellcolor{red!40}0.911 & \cellcolor{orange!40}0.914 & \cellcolor{orange!40}0.916\\
    SSS & \cellcolor{orange!40}0.898 & \cellcolor{orange!40}0.906 & \cellcolor{red!40}0.911 & \cellcolor{red!40}0.915 & \cellcolor{red!40}0.917\\
    \hline
    Scene - Components & \multicolumn{5}{c|}{Mip-NeRF 360 - garden}\\
    Method & 138k & 196k0 & 252k & 308k & 364k\\
    \hline
    3DGS & 0.366 & 0.386 & 0.399 & 0.412 & 0.424\\
    GES & 0.373 & 0.390 & 0.402 & 0.416 & 0.430\\
    3DHGS & \cellcolor{yellow!40}0.383 & \cellcolor{yellow!40}0.394 & \cellcolor{yellow!40}0.410 & \cellcolor{yellow!40}0.423 & \cellcolor{yellow!40}0.433\\
    3DGS-MCMC & \cellcolor{orange!40}0.757 & \cellcolor{orange!40}0.784 & \cellcolor{orange!40}0.802 & \cellcolor{orange!40}0.814 & \cellcolor{orange!40}0.823\\
    SSS & \cellcolor{red!40}0.758 & \cellcolor{red!40}0.785 & \cellcolor{red!40}0.803 & \cellcolor{red!40}0.815 & \cellcolor{red!40}0.825\\
    \hline
    Scene - Components & \multicolumn{5}{c|}{Mip-NeRF 360 - kitchen}\\
    Method & 241k & 336k & 432k & 528k & 624k\\
    \hline
    3DGS & 0.762 & 0.796 & 0.760 & 0.769 & 0.776\\
    GES & \cellcolor{yellow!40}0.856 & \cellcolor{yellow!40}0.838 & \cellcolor{yellow!40}0.875 & \cellcolor{yellow!40}0.908 & \cellcolor{yellow!40}0.912\\
    3DHGS & 0.811 & 0.801 & 0.698 & 0.706 & 0.734\\
    3DGS-MCMC &\cellcolor{orange!40} 0.919 & \cellcolor{orange!40}0.925 & \cellcolor{orange!40}0.928 & \cellcolor{orange!40}0.930 & \cellcolor{orange!40}0.932\\
    SSS & \cellcolor{red!40}0.921 & \cellcolor{red!40}0.926 & \cellcolor{red!40}0.929 & \cellcolor{red!40}0.932 & \cellcolor{red!40}0.934\\
    \hline
    Scene - Components & \multicolumn{5}{c|}{Mip-NeRF 360 - room}\\
    Method & 112k & 154k & 198k & 242k & 286k\\
    \hline
    3DGS & 0.720 & 0.769 & 0.791 & 0.813 & 0.815\\
    GES & \cellcolor{yellow!40}0.786 & \cellcolor{yellow!40}0.797 & \cellcolor{yellow!40}0.824 & \cellcolor{yellow!40}0.851 & \cellcolor{yellow!40}0.839\\
    3DHGS & 0.721 & 0.746 & 0.783 & 0.793 & 0.844\\
    3DGS-MCMC & \cellcolor{orange!40}0.911 & \cellcolor{orange!40}0.916 & \cellcolor{orange!40}0.920 & \cellcolor{orange!40}0.923 & \cellcolor{orange!40}0.925\\
    SSS & \cellcolor{red!40}0.912 & \cellcolor{red!40}0.918 & \cellcolor{red!40}0.923 & \cellcolor{red!40}0.926 & \cellcolor{red!40}0.928\\
    \hline
    Scene - Components & \multicolumn{5}{c|}{Mip-NeRF 360 - stump}\\
    Method & 32k & 44.8k & 57.6k & 70.4k & 83.2k\\
    \hline
    3DGS & \cellcolor{yellow!40}0.421 & \cellcolor{yellow!40}0.428 & \cellcolor{yellow!40}0.439 & \cellcolor{yellow!40}0.447 & \cellcolor{yellow!40}0.450\\
    GES & 0.391 & 0.391 & 0.384 & 0.376 & 0.371\\
    3DHGS & 0.404 & 0.414 & 0.424 & 0.423 & 0.424\\
    3DGS-MCMC & \cellcolor{red!40}0.594 &\cellcolor{red!40} 0.624 & \cellcolor{red!40}0.650 & \cellcolor{red!40}0.668 & \cellcolor{red!40}0.683\\
    SSS & \cellcolor{orange!40}0.588 & \cellcolor{orange!40}0.616 & \cellcolor{orange!40}0.638 & \cellcolor{orange!40}0.657 & \cellcolor{orange!40}0.670\\
\end{tabular}
}
\caption{
    \textbf{SSIM results of varying component numbers experiments for every scene in the Mip-NeRF 360 dataset.}
}
\label{tab:varying_360_ssim}
\end{table}

\begin{table}
\resizebox{0.47\textwidth}{!}{
\begin{tabular}{l|ccccccc}
    Scene - Components & \multicolumn{5}{c|}{Mip-NeRF 360 - average}\\
    Method & $\delta$ & $1.4\delta$ & $1.8\delta$ & $2.2\delta$ & $2.6\delta$\\
    \hline
    3DGS & 0.516 & 0.494 & 0.493 & 0.476 & 0.461\\
    GES & \cellcolor{yellow!40}0.485 & \cellcolor{yellow!40}0.466& \cellcolor{yellow!40}0.439 & \cellcolor{yellow!40}0.419 & \cellcolor{yellow!40}0.409\\
    3DHGS & 0.511 & 0.502 & 0.507 & 0.493 & 0.461\\
    3DGS-MCMC & \cellcolor{red!40}0.284 & \cellcolor{red!40}0.264 & \cellcolor{orange!40}0.250 & \cellcolor{orange!40}0.239 & \cellcolor{orange!40}0.231\\
    SSS & \cellcolor{red!40}0.284 & \cellcolor{red!40}0.264 & \cellcolor{red!40}0.249 & \cellcolor{red!40}0.237 & \cellcolor{red!40}0.228\\
    \hline
    Scene - Components & \multicolumn{5}{c|}{Mip-NeRF 360 - bicycle}\\
    Method & 54k & 75k & 97k & 118k & 140k\\
    \hline 
    3DGS & \cellcolor{yellow!40}0.640 & \cellcolor{yellow!40}0.637 & \cellcolor{yellow!40}0.632 & \cellcolor{yellow!40}0.631 & \cellcolor{yellow!40}0.636\\
    GES & \cellcolor{yellow!40}0.640 & 0.642 & 0.648 & 0.646 & 0.642\\
    3DHGS & 0.645 & 0.642 & 0.644 & 0.639 & 0.644\\
    3DGS-MCMC & \cellcolor{red!40}0.430 & \cellcolor{red!40}0.405& \cellcolor{red!40}0.387 & \cellcolor{red!40}0.371 & \cellcolor{red!40}0.360\\
    SSS & \cellcolor{orange!40}0.434 & \cellcolor{orange!40}0.412 & \cellcolor{orange!40}0.394 & \cellcolor{orange!40}0.379 & \cellcolor{orange!40}0.367\\
    \hline
    Scene - Components & \multicolumn{5}{c|}{Mip-NeRF 360 - bonsai}\\
    Method & 206k & 280k & 360k & 440k & 520k\\
    \hline
    3DGS & 0.388 & 0.360 & 0.361 & 0.352 & 0.340\\
    GES & \cellcolor{yellow!40}0.370 & \cellcolor{yellow!40}0.342 & \cellcolor{yellow!40}0.311 & \cellcolor{yellow!40}0.284 & \cellcolor{yellow!40}0.242\\
    3DHGS & 0.384 & 0.366& 0.360 & 0.347 & 0.343\\
    3DGS-MCMC & \cellcolor{orange!40}0.200 & \cellcolor{orange!40}0.189 & \cellcolor{orange!40}0.183 & \cellcolor{orange!40}0.178 & \cellcolor{orange!40}0.175\\
    SSS & \cellcolor{red!40}0.192 & \cellcolor{red!40}0.181 & \cellcolor{red!40}0.173 & \cellcolor{red!40}0.168 & \cellcolor{red!40}0.164\\
    \hline
    Scene - Components & \multicolumn{5}{c|}{Mip-NeRF 360 - counter}\\
    Method & 155k & 224k & 288k & 352k & 416k\\
    \hline
    3DGS & 0.499 & 0.476 & 0.477 & 0.427 & 0.352\\
    GES & \cellcolor{yellow!40}0.461 & 0\cellcolor{yellow!40}.359 & \cellcolor{yellow!40}0.284 & \cellcolor{yellow!40}0.266 & \cellcolor{yellow!40}0.242\\
    3DHGS & 0.513 & 0.495 & 0.480 & 0.438 & 0.364\\
    3DGS-MCMC & \cellcolor{red!40}0.212 & \cellcolor{orange!40}0.200 & \cellcolor{orange!40}0.192 & \cellcolor{orange!40}0.187 & \cellcolor{orange!40}0.183\\
    SSS & \cellcolor{orange!40}0.214 & \cellcolor{red!40}0.197 & \cellcolor{red!40}0.188 & \cellcolor{red!40}0.181 & \cellcolor{red!40}0.176\\
    \hline
    Scene - Components & \multicolumn{5}{c|}{Mip-NeRF 360 - garden}\\
    Method & 138k & 196k & 252k & 308k & 364k\\
    \hline
    3DGS & 0.661 & 0.640  & 0.627 & 0.613 & 0.601\\
    GES & 0.652 & \cellcolor{yellow!40}0.635 & \cellcolor{yellow!40}0.623 & 0.607 & 0.594\\
    3DHGS & \cellcolor{yellow!40}0.649 & 0.638 & \cellcolor{yellow!40}0.623 & \cellcolor{yellow!40}0.606 & \cellcolor{yellow!40}0.541\\
    3DGS-MCMC & \cellcolor{red!40}0.296 & \cellcolor{red!40}0.256 & \cellcolor{orange!40}0.229 & \cellcolor{orange!40}0.209 & \cellcolor{orange!40}0.194\\
    SSS & \cellcolor{orange!40}0.297 & \cellcolor{orange!40}0.257 & \cellcolor{red!40}0.228 & \cellcolor{red!40}0.207 & \cellcolor{red!40}0.191\\
    \hline
    Scene - Components & \multicolumn{5}{c|}{Mip-NeRF 360 - kitchen}\\
    Method & 241k & 336k & 432k & 528k & 624k\\
    \hline
    3DGS & 0.347 & 0.312 & 0.353 & 0.344 & 0.334\\
    GES & \cellcolor{yellow!40}0.235 & \cellcolor{yellow!40}0.260 & \cellcolor{yellow!40}0.211 & \cellcolor{yellow!40}0.163 & \cellcolor{yellow!40}0.156\\
    3DHGS & 0.295 & 0.308 & 0.413 & 0.406 & 0.378\\
    3DGS-MCMC & \cellcolor{orange!40}0.145 & \cellcolor{orange!40}0.134 & \cellcolor{orange!40}0.129 & \cellcolor{orange!40}0.125 & \cellcolor{orange!40}0.122\\
    SSS & \cellcolor{red!40}0.142 & \cellcolor{red!40}0.132 & \cellcolor{red!40}0.125 & \cellcolor{red!40}0.119 & \cellcolor{red!40}0.115\\
    \hline
    Scene - Components & \multicolumn{5}{c|}{Mip-NeRF 360 - room}\\
    Method & 112k & 154k & 198k & 242k & 286k\\
    \hline
    3DGS & 0.471 & 0.429 & 0.403 & 0.376 & 0.375\\
    GES & \cellcolor{yellow!40}0.404 & \cellcolor{yellow!40}0.396 & \cellcolor{yellow!40}0.361 & \cellcolor{yellow!40}0.325 & \cellcolor{yellow!40}0.342\\
    3DHGS & 0.467 & 0.449 & 0.413 & 0.401 & 0.343\\
    3DGS-MCMC & \cellcolor{orange!40}0.235 & \cellcolor{orange!40}0.223 & \cellcolor{orange!40}0.215 & \cellcolor{orange!40}0.209 & \cellcolor{orange!40}0.203\\
    SSS & \cellcolor{red!40}0.230 & \cellcolor{red!40}0.217 & \cellcolor{red!40}0.207 & \cellcolor{red!40}0.200 & \cellcolor{red!40}0.195\\
    \hline
    Scene - Components & \multicolumn{5}{c|}{Mip-NeRF 360 - stump}\\
    Method & 32k & 44.8k & 57.6k & 70.4k & 83.2k\\
    \hline
    3DGS & 0.610 & 0.605 & 0.595 & 0.590 & 0.588\\
    GES & 0.634 & 0.632 & 0.639 & 0.642 & 0.647\\
    3DHGS & 0.627 & 0.619 & 0.612 & 0.612 & 0.613\\
    3DGS-MCMC & \cellcolor{red!40}0.474 & \cellcolor{red!40}0.442 & \cellcolor{red!40}0.415 & \cellcolor{red!40}0.395 & \cellcolor{red!40}0.379\\
    SSS & \cellcolor{orange!40}0.481 & \cellcolor{orange!40}0.451 & \cellcolor{orange!40}0.427 & \cellcolor{orange!40}0.406 & \cellcolor{orange!40}0.390\\
\end{tabular}
}
\caption{
    \textbf{LPIPS results of varying component numbers experiments for every scene in the Mip-NeRF 360 dataset.}
}
\label{tab:varying_360_lpips}
\end{table}

\begin{table}
\resizebox{0.48\textwidth}{!}{
\begin{tabular}{l|ccccccc}
    Scene - Components & \multicolumn{5}{c|}{Tanks\&Temples - average}\\
    Method & $\delta$ & $1.4\delta$ & $1.8\delta$ & $2.2\delta$ & $2.6\delta$\\
    \hline
    3DGS & 15.70 & 15.79 & 15.88 & 16.37 & 16.64\\
    GES & \cellcolor{yellow!40}16.74 & \cellcolor{yellow!40}17.47 & \cellcolor{yellow!40}17.80 & \cellcolor{yellow!40}18.38 & \cellcolor{yellow!40}18.71\\
    3DHGS & 15.13 & 15.26 & 15.36 & 15.55 & 15.69\\
    3DGS-MCMC & \cellcolor{orange!40}23.18 & \cellcolor{orange!40}23.53 & \cellcolor{orange!40}23.65 & \cellcolor{orange!40}23.84 & \cellcolor{orange!40}23.93\\
    SSS & \cellcolor{red!40}23.67 & \cellcolor{red!40}24.03 & \cellcolor{red!40}24.19 & \cellcolor{red!40}24.35 & \cellcolor{red!40}24.40\\
    \hline
    Scene - Components & \multicolumn{5}{c|}{Tanks\&Temples - train}\\
    Method & 182k & 252k & 324k & 396k & 468k\\
    \hline 
    3DGS & 15.20 & 15.53 & 15.58 & 16.45 & 16.92\\
    GES & \cellcolor{yellow!40}18.15 & \cellcolor{yellow!40}19.39 & \cellcolor{yellow!40}19.94 & \cellcolor{yellow!40}21.14 & \cellcolor{yellow!40}21.58\\
    3DHGS & 14.42 & 14.62 & 14.75 & 15.05 & 15.24\\
    3DGS-MCMC & \cellcolor{orange!40}21.72 & \cellcolor{orange!40}22.07 & \cellcolor{orange!40}22.03 & \cellcolor{orange!40}22.24 & \cellcolor{orange!40}22.34\\
    SSS & \cellcolor{red!40}22.46 & \cellcolor{red!40}22.74 & \cellcolor{red!40}22.84 & \cellcolor{red!40}22.92 & \cellcolor{red!40}22.97\\
    \hline
    Scene - Components & \multicolumn{5}{c|}{Tanks\&Temples - truck}\\
    Method & 136k & 196k & 252k & 308k & 364k\\
    \hline
    3DGS & \cellcolor{yellow!40}16.20 & \cellcolor{yellow!40}16.06 & \cellcolor{yellow!40}16.18 & \cellcolor{yellow!40}16.29 & \cellcolor{yellow!40}16.36\\
    GES & 15.32 & 15.54 & 15.66 & 15.62 & 15.85\\
    3DHGS & 15.84 & 15.90 & 15.97 & 16.05 & 16.13\\
    3DGS-MCMC & \cellcolor{orange!40}24.64 & \cellcolor{orange!40}24.99 & \cellcolor{orange!40}25.27 & \cellcolor{orange!40}25.44 & \cellcolor{orange!40}25.53\\
    SSS & \cellcolor{red!40}24.88 & \cellcolor{red!40}25.33 & \cellcolor{red!40}25.53 & \cellcolor{red!40}25.74 & \cellcolor{red!40}25.88\\
    \hline
    Scene - Components & \multicolumn{5}{c|}{Deep Blending - average}\\
    Method & $\delta$ & $1.4\delta$ & $1.8\delta$ & $2.2\delta$ & $2.6\delta$\\
    \hline
    3DGS & 15.95 & 16.05 & 16.33 & 16.60 & 17.06\\
    GES & 15.99 & 16.10 & 16.55 & 17.38 & 17.23\\
    3DHGS & \cellcolor{yellow!40}16.22 & \cellcolor{yellow!40}16.63 & \cellcolor{yellow!40}16.54 & \cellcolor{yellow!40}18.02 & \cellcolor{yellow!40}18.07\\
    3DGS-MCMC & \cellcolor{orange!40}28.38 & \cellcolor{orange!40}28.82 & \cellcolor{orange!40}29.11 & \cellcolor{orange!40}29.17 & \cellcolor{orange!40}29.35\\
    SSS & \cellcolor{red!40}28.69 & \cellcolor{red!40}29.08 & \cellcolor{red!40}29.29 & \cellcolor{red!40}29.41 & \cellcolor{red!40}29.67\\
    \hline
    Scene - Components & \multicolumn{5}{c|}{Deep Blending - drjohnson}\\
    Method & 80k & 112k & 144k & 176k & 208k\\
    \hline
    3DGS & 16.66 & 16.64 & 17.05 & 16.91 & 17.34\\
    GES & 16.43 & 16.84 & 17.13 & 17.41 & 18.05\\
    3DHGS & \cellcolor{yellow!40}17.17 & \cellcolor{yellow!40}17.29 & \cellcolor{yellow!40}17.37 & \cellcolor{yellow!40}19.19 & \cellcolor{yellow!40}19.82\\
    3DGS-MCMC & \cellcolor{orange!40}28.15 & \cellcolor{orange!40}28.62 & \cellcolor{orange!40}28.83 & \cellcolor{orange!40}28.90 & \cellcolor{orange!40}28.94\\
    SSS & \cellcolor{red!40}28.79 & \cellcolor{red!40}29.05 & \cellcolor{red!40}29.19 & \cellcolor{red!40}29.43 & \cellcolor{red!40}29.42\\
    \hline
    Scene - Components & \multicolumn{5}{c|}{Deep Blending - playroom}\\
    Method & 37k & 51.8k & 66.6k & 81.4k & 96.2k\\
    \hline
    3DGS & 15.23 & 15.45 & 15.61 & 16.28 & 16.78\\
    GES & \cellcolor{yellow!40}15.56 & 15.36 & \cellcolor{yellow!40}15.96 & \cellcolor{yellow!40}17.34 & 16.42\\
    3DHGS & 15.26 & \cellcolor{yellow!40}15.96 & 15.71 & 16.22 & \cellcolor{yellow!40}16.94\\
    3DGS-MCMC & \cellcolor{red!40}28.61 & \cellcolor{orange!40}29.02 & \cellcolor{orange!40}29.39 & \cellcolor{red!40}29.45 & \cellcolor{orange!40}29.76\\
    SSS & \cellcolor{orange!40}28.58 & \cellcolor{red!40}29.10 & \cellcolor{red!40}29.40 & \cellcolor{orange!40}29.40 & \cellcolor{red!40}29.91\\
\end{tabular}
}
\caption{
    \textbf{PSNR results of varying component numbers experiments for every scene in Tanks\&Temples and Deep Blending dataset.}
}
\label{tab:varying_td_psnr}
\end{table}

\begin{table}
\resizebox{0.48\textwidth}{!}{
\begin{tabular}{l|ccccccc}
    Scene - Components & \multicolumn{5}{c|}{Tanks\&Temples - average}\\
    Method & $\delta$ & $1.4\delta$ & $1.8\delta$ & $2.2\delta$ & $2.6\delta$\\
    \hline
    3DGS & 0.528 & 0.537 & 0.543 & 0.566 & 0.580\\
    GES & \cellcolor{yellow!40}0.589 & \cellcolor{yellow!40}0.623 & \cellcolor{yellow!40}0.642 & \cellcolor{yellow!40}0.668 & \cellcolor{yellow!40}0.680\\
    3DHGS & 0.504 & 0.511 & 0.518 & 0.526 & 0.533\\
    3DGS-MCMC & \cellcolor{orange!40}0.826 & \cellcolor{orange!40}0.836 & \cellcolor{orange!40}0.842 & \cellcolor{orange!40}0.848 & \cellcolor{orange!40}0.851\\
    SSS & \cellcolor{red!40}0.827 & \cellcolor{red!40}0.839 & \cellcolor{red!40}0.846 & \cellcolor{red!40}0.852 & \cellcolor{red!40}0.856\\
    \hline
    Scene - Components & \multicolumn{5}{c|}{Tanks\&Temples - train}\\
    Method & 182k & 252k & 324k & 396k & 468k\\
    \hline 
    3DGS & 0.506 & 0.521 & 0.523 & 0.561 & 0.584\\
    GES & \cellcolor{yellow!40}0.645 & \cellcolor{yellow!40}0.699 & \cellcolor{yellow!40}0.727 & \cellcolor{yellow!40}0.774 & \cellcolor{yellow!40}0.788\\
    3DHGS & 0.473 & 0.479 & 0.485 & 0.494 & 0.505\\
    3DGS-MCMC & \cellcolor{orange!40}0.798 & \cellcolor{orange!40}0.807 & \cellcolor{orange!40}0.814 & \cellcolor{orange!40}0.820 & \cellcolor{orange!40}0.824\\
    SSS & \cellcolor{red!40}0.799 & \cellcolor{red!40}0.812 & \cellcolor{red!40}0.821 & \cellcolor{red!40}0.828 & \cellcolor{red!40}0.833\\
    \hline
    Scene - Components & \multicolumn{5}{c|}{Tanks\&Temples - truck}\\
    Method & 136k & 196k & 252k & 308k & 364k\\
    \hline
    3DGS & \cellcolor{yellow!40}0.549 & \cellcolor{yellow!40}0.554 & \cellcolor{yellow!40}0.563 & \cellcolor{yellow!40}0.571 & \cellcolor{yellow!40}0.576\\
    GES & 0.532 & 0.548 & 0.557 & 0.562 & 0.572\\
    3DHGS & 0.535 & 0.543 & 0.550 & 0.557 & 0.561\\
    3DGS-MCMC & \cellcolor{red!40}0.855 & \cellcolor{red!40}0.865 & \cellcolor{orange!40}0.871 & \cellcolor{orange!40}0.876 & \cellcolor{orange!40}0.879\\
    SSS & \cellcolor{orange!40}0.854 & \cellcolor{red!40}0.865 & \cellcolor{red!40}0.872 & \cellcolor{red!40}0.877 & \cellcolor{red!40}0.880\\
    \hline
    Scene - Components & \multicolumn{5}{c|}{Deep Blending - average}\\
    Method & $\delta$ & $1.4\delta$ & $1.8\delta$ & $2.2\delta$ & $2.6\delta$\\
    \hline
    3DGS & 0.702 & 0.706 & 0.711 & 0.715 & 0.726\\
    GES & \cellcolor{yellow!40}0.704 & 0.707 & \cellcolor{yellow!40}0.717 & 0.732 & 0.732\\
    3DHGS & \cellcolor{yellow!40}0.704 & \cellcolor{yellow!40}0.713 & 0.710 & \cellcolor{yellow!40}0.737 & \cellcolor{yellow!40}0.737\\
    3DGS-MCMC & \cellcolor{red!40}0.881 & \cellcolor{red!40}0.887 & \cellcolor{orange!40}0.890 & \cellcolor{red!40}0.893 & \cellcolor{orange!40}0.895\\
    SSS & \cellcolor{red!40}0.881 & \cellcolor{red!40}0.887 & \cellcolor{red!40}0.891 & \cellcolor{red!40}0.893 & \cellcolor{red!40}0.897\\
    \hline
    Scene - Components & \multicolumn{5}{c|}{Deep Blending - drjohnson}\\
    Method & 80k & 112k & 144k & 176k & 208k\\
    \hline
    3DGS & 0.698 & 0.701 & 0.707 & 0.707 & 0.717\\
    GES & 0.696 & 0.705 & \cellcolor{yellow!40}0.715 & 0.721 & 0.735\\
    3DHGS & \cellcolor{yellow!40}0.706 & \cellcolor{yellow!40}0.711 & 0.711 & \cellcolor{yellow!40}0.752 & \cellcolor{yellow!40}0.744\\
    3DGS-MCMC & \cellcolor{red!40}0.880 & \cellcolor{red!40}0.885 & \cellcolor{orange!40}0.888 & \cellcolor{orange!40}0.890 & \cellcolor{orange!40}0.891\\
    SSS & \cellcolor{red!40}0.880 & \cellcolor{red!40}0.885 & \cellcolor{red!40}0.890 & \cellcolor{red!40}0.894 & \cellcolor{red!40}0.896\\
    \hline
    Scene - Components & \multicolumn{5}{c|}{Deep Blending - playroom}\\
    Method & 37k & 51.8k & 66.6k & 81.4k & 96.2k\\
    \hline
    3DGS & 0.706 & 0.711 & 0.714 & 0.723 & 0.735\\
    GES & 0.713 & 0.709 & 0.719 & 0.742 & 0.730\\
    3DHGS & 0.701 & 0.714 & 0.709 & 0.722 & 0.731\\
    3DGS-MCMC & \cellcolor{red!40}0.883 & \cellcolor{red!40}0.890 & \cellcolor{red!40}0.893 & \cellcolor{red!40}0.896 & \cellcolor{red!40}0.899\\
    SSS & \cellcolor{orange!40}0.882 & \cellcolor{orange!40}0.888 & \cellcolor{orange!40}0.892 & \cellcolor{orange!40}0.891 & \cellcolor{orange!40}0.897\\
\end{tabular}
}
\caption{
    \textbf{SSIM results of varying component numbers experiments for every scene in Tanks\&Temples and Deep Blending dataset.}
}
\label{tab:varying_td_ssim}
\end{table}

\begin{table}
\resizebox{0.48\textwidth}{!}{
\begin{tabular}{l|ccccccc}
Scene - Components & \multicolumn{5}{c|}{Tanks\&Temples - average}\\
    Method & $\delta$ & $1.4\delta$ & $1.8\delta$ & $2.2\delta$ & $2.6\delta$\\
    \hline
    3DGS & 0.569 & 0.558& 0.552 & 0.527 & 0.510\\
    GES & \cellcolor{yellow!40}0.500 & \cellcolor{yellow!40}0.460 & \cellcolor{yellow!40}0.441 & \cellcolor{yellow!40}0.410 & \cellcolor{yellow!40}0.397\\
    3DHGS & 0.590 & 0.582 & 0.574 & 0.566 & 0.558\\
    3DGS-MCMC & \cellcolor{orange!40}0.231 & \cellcolor{orange!40}0.214 & \cellcolor{orange!40}0.203 & \cellcolor{orange!40}0.195 & \cellcolor{orange!40}0.187\\
    SSS & \cellcolor{red!40}0.229 & \cellcolor{red!40}0.210 & \cellcolor{red!40}0.195 & \cellcolor{red!40}0.185 & \cellcolor{red!40}0.177\\
    \hline
    Scene - Components & \multicolumn{5}{c|}{Tanks\&Temples - train}\\
    Method & 182k & 252k & 324k & 396k & 468k\\
    \hline 
    3DGS & 0.564 & 0.548 & 0.543 & 0.504 & 0.476\\
    GES & \cellcolor{yellow!40}0.415 & \cellcolor{yellow!40}0.352 & \cellcolor{yellow!40}0.321 & \cellcolor{yellow!40}0.267 & \cellcolor{yellow!40}0.251\\
    3DHGS & 0.597 & 0.592 & 0.583 & 0.575 & 0.561\\
    3DGS-MCMC & \cellcolor{orange!40}0.255 & \cellcolor{orange!40}0.240 & \cellcolor{orange!40}0.229 & \cellcolor{orange!40}0.220 & \cellcolor{orange!40}0.212\\
    SSS & \cellcolor{red!40}0.254 & \cellcolor{red!40}0.235 & \cellcolor{red!40}0.220 & \cellcolor{red!40}0.208 & \cellcolor{red!40}0.199\\
    \hline
    Scene - Components & \multicolumn{5}{c|}{Tanks\&Temples - truck}\\
    Method & 136k & 196k & 252k & 308k & 364k\\
    \hline
    3DGS & \cellcolor{yellow!40}0.575 & \cellcolor{yellow!40}0.569 & \cellcolor{yellow!40}0.560 & \cellcolor{yellow!40}0.551 & 0.545\\
    GES & 0.585 & \cellcolor{yellow!40}0.569 & 0.561 & 0.552 & \cellcolor{yellow!40}0.543\\
    3DHGS & 0.583 & 0.571 & 0.564& 0.558 & 0.555\\
    3DGS-MCMC & \cellcolor{orange!40}0.206 & \cellcolor{orange!40}0.189 & \cellcolor{orange!40}0.178 & \cellcolor{orange!40}0.170 & \cellcolor{orange!40}0.163\\
    SSS & \cellcolor{red!40}0.204 & \cellcolor{red!40}0.184& \cellcolor{red!40}0.170 & \cellcolor{red!40}0.161 & \cellcolor{red!40}0.154\\
    \hline
    Scene - Components & \multicolumn{5}{c|}{Deep Blending - average}\\
    Method & $\delta$ & $1.4\delta$ & $1.8\delta$ & $2.2\delta$ & $2.6\delta$\\
    \hline
    3DGS & 0.525 & 0.522 & 0.515 & 0.510 & 0.500\\
    GES & \cellcolor{yellow!40}0.522 & 0.522 & \cellcolor{yellow!40}0.509 & \cellcolor{yellow!40}0.495 & 0.491\\
    3DHGS & \cellcolor{yellow!40}0.522 & \cellcolor{yellow!40}0.515 & 0.515 & 0.490 & \cellcolor{yellow!40}0.488\\
    3DGS-MCMC & \cellcolor{orange!40}0.325 & \cellcolor{orange!40}0.310 & \cellcolor{orange!40}0.301 & \cellcolor{orange!40}0.295 & \cellcolor{orange!40}0.290\\
    SSS & \cellcolor{red!40}0.319 & \cellcolor{red!40}0.305 & \cellcolor{red!40}0.296 & \cellcolor{red!40}0.289 & \cellcolor{red!40}0.282\\
    \hline
    Scene - Components & \multicolumn{5}{c|}{Deep Blending - drjohnson}\\
    Method & 80k & 112k & 144k & 176k & 208k\\
    \hline
    3DGS & 0.526 & 0.523 & 0.514 & 0.512 & 0.504\\
    GES & 0.529 & 0.522 & \cellcolor{yellow!40}0.508 & 0.503 & 0.487\\
    3DHGS & \cellcolor{yellow!40}0.515 & \cellcolor{yellow!40}0.515 & 0.511 & \cellcolor{yellow!40}0.476 & \cellcolor{yellow!40}0.482\\
    3DGS-MCMC & \cellcolor{orange!40}0.316 &\cellcolor{red!40} 0.290 & \cellcolor{orange!40}0.290 & \cellcolor{orange!40}0.284 & \cellcolor{orange!40}0.280\\
    SSS & \cellcolor{red!40}0.310 & \cellcolor{orange!40}0.297 & \cellcolor{red!40}0.288 &\cellcolor{red!40} 0.280 & \cellcolor{red!40}0.275\\
    \hline
    Scene - Components & \multicolumn{5}{c|}{Deep Blending - playroom}\\
    Method & 37k & 51.8k & 66.6k & 81.4k & 96.2k\\
    \hline
    3DGS & 0.524 & 0.521 & 0.515 & 0.508 & 0.496\\
    GES & \cellcolor{yellow!40}0.516 & 0.522 & \cellcolor{yellow!40}0.509 & \cellcolor{yellow!40}0.488 & \cellcolor{yellow!40}0.494\\
    3DHGS & 0.530 & \cellcolor{yellow!40}0.515 & 0.519 & 0.505 & \cellcolor{yellow!40}0.494\\
    3DGS-MCMC & \cellcolor{orange!40}0.333 & \cellcolor{orange!40}0.320 & \cellcolor{orange!40}0.312 & \cellcolor{orange!40}0.306 & \cellcolor{orange!40}0.301\\
    SSS & \cellcolor{red!40}0.329 & \cellcolor{red!40}0.314 & \cellcolor{red!40}0.304 & \cellcolor{red!40}0.298 & \cellcolor{red!40}0.289\\
\end{tabular}
}
\caption{
    \textbf{LPIPS results of varying component numbers experiments for every scene in Tanks\&Temples and Deep Blending dataset.}
}
\label{tab:varying_td_lpips}
\end{table}

\subsection{Ablation Study}
\label{sec:ablation_results}

\begin{table}
\begin{tabular}{l|ccc|}
    Ablation Setup | Metric & PSNR $\uparrow$   & SSIM$\uparrow$   & LPIPS$\downarrow$\\
    \hline
    Mip-NeRF & 22.22 & 0.759 & 0.257\\
    3DGS & 23.14 & 0.841 & 0.183\\
    GES & 23.35 & 0.836 & 0.198\\
    \hline 
    SGD + positive t-dis &  23.80 & 0.838 & 0.191\\
    SGHMC + positive t-dis & 24.53 & 0.864 & 0.155\\
    Full model & 24.87 & 0.873 & 0.138
\end{tabular}
\caption{
    \textbf{Ablation Study on Tanks\&Temples with the same component numbers as in baselines. More details are in the SM.}
}
\label{tab:ablation}
\end{table}
We conduct an ablation study to show the effectiveness of various components in SSS. We report the results on one dataset in \cref{tab:ablation}. Universally, by only replacing Gaussians with Student's t distributions (SGD+t-distribution), it already outperforms Mip-NeRF, 3DGS, and GES, demonstrating improved expressivity. Further with SGHMC, it is already the best method, showing the advantage of the proposed sampling and the importance of a good sampler in the optimization process. Finally, adding negative components further improves the results.

The detailed results of Ablation Study (effect of each contribution in SSS) on every scene with PSNR, SSIM, and LPIPS metrics among Mip-NeRF 360, Tanks \& Temples and Deep Blending datasets are in~\cref{tab:ablation_360_psnr,tab:ablation_360_ssim,tab:ablation_360_lpips,tab:ablation_td_psnr,tab:ablation_td_ssim,tab:ablation_td_lpips}.


\begin{table}[h]
\resizebox{0.48\textwidth}{!}{
\begin{tabular}{l|ccccccc|c}
    Method $\backslash$ Scene & bicycle & bonsai & counter & garden & kitchen & room & stump & average\\
    \hline 
    SGD + positive t-dis  & 25.52 & 31.80 & 28.82 & 27.60 & 30.93 & 31.60 & 26.72 & 29.00 \\
    SGHMC + positive t-dis & 25.97 & 32.87 & 29.49 & 27.92 & 31.92 & 32.04 & 27.44 & 29.66\\
    Full SSS model & 25.68 & 33.50 & 29.87 & 28.09 & 32.43 & 32.57 & 27.17 & 29.90\\
\end{tabular}
}
\caption{
    \textbf{PSNR results of ablation study for every scene in Mip-NeRF 360 dataset.}
}
\label{tab:ablation_360_psnr}
\end{table}

\begin{table}[h]
\resizebox{0.48\textwidth}{!}{
\begin{tabular}{l|ccccccc|c}
    Method $\backslash$ Scene & bicycle & bonsai & counter & garden & kitchen & room & stump & average\\
    \hline 
    SGD + positive t-dis & 0.767& 0.941 & 0.909 & 0.866 & 0.927 & 0.923 & 0.771 & 0.872\\
    SGHMC + positive t-dis & 0.801 & 0.952 & 0.920 & 0.879 & 0.936 & 0.933 & 0.817 & 0.891\\
    Full SSS model & 0.798 & 0.956 & 0.926 & 0.882 & 0.939 & 0.938 & 0.813 & 0.893\\
\end{tabular}
}
\caption{
    \textbf{SSIM results of ablation study for every scene in Mip-NeRF 360 dataset.}
}
\label{tab:ablation_360_ssim}
\end{table}

\begin{table}[h]
\resizebox{0.48\textwidth}{!}{
\begin{tabular}{l|ccccccc|c}
    Method $\backslash$ Scene & bicycle & bonsai & counter & garden & kitchen & room & stump & average\\
    \hline 
    SGD + positive t-dis & 0.225 & 0.185 & 0.188 & 0.110 & 0.121 & 0.199 & 0.234 & 0.180\\
    SGHMC + positive t-dis &  0.182 & 0.159 & 0.168 & 0.099 & 0.112 & 0.181 & 0.183 & 0.155\\
    Full SSS model & 0.173 & 0.151 & 0.156 & 0.009 & 0.104 & 0.167 & 0.174 & 0.145\\
\end{tabular}
}
\caption{
    \textbf{LPIPS results of ablation study for every scene in Mip-NeRF 360 dataset.}
}
\label{tab:ablation_360_lpips}
\end{table}

\begin{table}[h]
\resizebox{0.48\textwidth}{!}{
\begin{tabular}{l|ccccccc}
    Dataset - Scene & \multicolumn{3}{c|}{Tanks\&Temples} & \multicolumn{3}{c}{Deep Blending}\\
    Method & train & truck & average & drjohnson & playroom & average\\
    \hline 
    SGD + positive t-dis & 22.18 & 25.42 & 23.80 & 29.22 & 29.93 & 29.57\\
    SGHMC + positive t-dis & 22.92 & 26.15 & 24.53 & 29.45 & 30.04 & 29.75\\
    Full SSS model &23.32 & 26.41 & 24.87 & 29.66 & 30.47 & 30.07\\
\end{tabular}
}
\caption{
    \textbf{PSNR results of ablation study for every scene in Tanks\&Temples and Deep Blending dataset.}
}
\label{tab:ablation_td_psnr}
\end{table}

\begin{table}[h]
\resizebox{0.48\textwidth}{!}{
\begin{tabular}{l|ccccccc}
    Dataset - Scene & \multicolumn{3}{c|}{Tanks\&Temples} & \multicolumn{3}{c}{Deep Blending}\\
    Method & train & truck & average & drjohnson & playroom & average\\
    \hline 
    SGD + positive t-dis & 0.803 & 0.874 & 0.838 & 0.900 & 0.901 & 0.901\\
    SGHMC + positive t-dis & 0.838 & 0.891 & 0.864 & 0.902 & 0.905 & 0.903\\
    Full SSS model & 0.850 & 0.897 & 0.873 & 0.905 & 0.909 & 0.907\\
\end{tabular}
}
\caption{
    \textbf{SSIM results of ablation study for every scene in Tanks\&Temples and Deep Blending dataset.}
}
\label{tab:ablation_td_ssim}
\end{table}

\begin{table}[h]
\resizebox{0.48\textwidth}{!}{
\begin{tabular}{l|ccccccc}
    Dataset - Scene & \multicolumn{3}{c|}{Tanks\&Temples} & \multicolumn{3}{c}{Deep Blending}\\
    Method & train & truck & average & drjohnson & playroom & average\\
    \hline 
    SGD + positive t-dis & 0.226 & 0.156 & 0.191 & 0.248 & 0.247 & 0.247\\
    SGHMC + positive t-dis & 0.186 & 0.124 & 0.155 & 0.262 & 0.257 & 0.260\\
    Full SSS model & 0.166 & 0.109 & 0.138 & 0.249 & 0.245 & 0.247\\
\end{tabular}
}
\caption{
    \textbf{LPIPS results of ablation study for every scene in Tanks\&Temples and Deep Blending dataset.}
}
\label{tab:ablation_td_lpips}
\end{table}

\paragraph{More ablation} We also show comparison of applying SGHMC with vanilla 3DGS and positive t-distributions only with Tanks \& Temples and Deep Blending datasets to the ablations (\cref{tab:extra_ablation}). Replacing SGD with SGHMC already improves the results. Replacing Gaussians with positive t-distributions further improves the PSNR but slightly reduces SSIM and LPIPS. Nonetheless, our Full model is obviously the best. While individual techniques alone might provide merely small improvements, SSS as a whole is the SOTA.

\begin{table}[tb]
\resizebox{0.48\textwidth}{!}{
\begin{tabular}{l|ccc|ccc}
    Dataset  & \multicolumn{3}{c|}{Tanks\&Temples} & \multicolumn{3}{c}{Deep Blending}\\
    Method|Metric & PSNR $\uparrow$   & SSIM$\uparrow$   & LPIPS$\downarrow$ & PSNR $\uparrow$   & SSIM$\uparrow$   & LPIPS$\downarrow$\\
    \hline
    3DGS & 23.14 & 0.841  & 0.183 & 29.41  & 0.903  & 0.243 \\
    SGD + positive t-dis & 23.80  & 0.838  & 0.191 & 29.57 & 0.901 & 0.247 \\
    \hline
    SGHMC + 3DGS & 24.52 & 0.869 & 0.150 & 29.56 & 0.906 & 0.239 \\
    SGHMC + positive t-dis & 24.53 & 0.864  & 0.155 & 29.75  & 0.903 & 0.259 \\
    \hline
    Full model & 24.87 & 0.873 & 0.138  & 30.07 & 0.907 & 0.247 \\
\end{tabular}
}
\caption{
    \textbf{More Ablation Study.} 
}
\label{tab:extra_ablation}
\end{table}

\section{Sampling Effects on Learning}
When trying the original SGD with only positive Student's t components, the learned $\nu$ values were not ideal. Considering Gaussian is simply a Student's t distribution with fixed $\nu=\infty$, it shows that $\nu$ introduces undesirable local minima. This was mitigated by SGHMC as the friction term decouples parameters, but the sampling became slow compared to vanilla 3DGS.

\begin{figure}
    \centering
    \includegraphics[width=0.8\columnwidth]{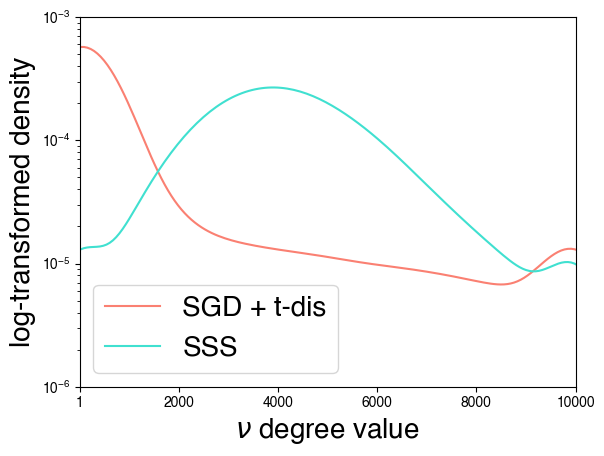}
    \caption{\textbf{Sampling effects on learning} .}
    \label{fig:nu}
\end{figure}

We compare the learned $\nu$ distributions between SGHMC (decoupling) and the 3DGS optimization (no decoupling) in~\cref{fig:nu}. SGHMC learned a distribution across a wide range, with no mode collapse and fully utilizing the representation power of t-distribution. 3DGS optimization learns a distribution heavily concentrated in some areas (near 1, likely mode collapse), unable to explore the full space of t-distribution.

\section{Implementation Details}
The implementation of SSS is based on open-source codes of the vanilla 3D Gaussian Splatting (3DGS)~\cite{kerbl20233d} and 3DGS-MCMC~\cite{kheradmand20243d}. We modified various parts to replace Gaussians with Student's t distributions for splatting algorithm~\cite{zwicker2001ewa, zwicker2002ewa}. These modifications are adapted to both forward and backward procedures. We show the formulae of the forward process (transformation and marginalization) in \cref{sec:forward_pass}. For backward propagation with training, we derive the relevant partial derivatives with respect to Student's t distribution, which is given in \cref{sec:backward}. Theoretically, the value of $\nu$ of Student's t distribution can be infinite, but in practice, to avoid numerical issues, we limit the range of $\nu$ from 1 to 10000.

There are two major differences between using Student's t distribution and Gaussian distribution. The first is that a t-distribution is not a t-distribution anymore after convolution with another t-distribution. So, we did not add a low-pass filter like in vanilla 3DGS (adding $0.3$ to the diagonal values of the projected covariance matrix) in our final model. However, we do add the low-pass filter as a practical solution in SGD + positive t-distribution (replacing Gaussian with Student's t but without negative components and SGHMC) in the Ablation Study for fair comparison with similar component numbers. This will be discussed in~\cref{sec:ablation_implementation}. Another difference is that Student's t distribution does not have a general ``empirical rule" as Gaussian distribution. In vanilla 3DGS, they use the ``empirical rule" (also known as ``68–95–99.7 rule" or ``three-sigma rule") to truncate the projected Gaussian in 2D space. The ``three-sigma rule" can be applied to t-distributions with thin tails (high $\nu$ degree). Besides, we obtain different critical values of different $\nu$ degrees~\cite{heckert2002handbook} and interpolate these to obtain appropriate truncated values for fat-tailed t-distributions with lower $\nu$ values.

Our training process does not use the adaptive density control in vanilla 3DGS. Instead, we recycle components with low opacity to high opacity components every $n$ iterations. This is done by creating a Multinomial distribution of opacity values and sampling all components that have relatively high probability. Recycle can only be achieved if certain conditions are met, \ie the change to the current state (rendering result) after the recycle is minimal~\cite{rota2024revising, kheradmand20243d}. We give detailed formulae of recycling in \cref{sec:relocation}. The result of calculating the new covariance matrix after recycling includes the $\beta()$ function. Because CUDA native implementation does not include the support for the $\beta()$ function, we decompose the $\beta()$ function into $\Gamma()$ function for computation. Further, to prevent excessive values by the $\Gamma()$ function with a large input number, we further use the $\ln(\Gamma())$ function in practice. These are shown with~\cref{eq:beta3,eq:beta4}.

Finally, we employ the Adam gradient in SGHMC. The hyperparameters in our training process are largely the same as those of the original 3DGS and 3DGS-MCMC (\eg we down-scale images to the same resolution for large scenes in the Mip-NeRF360 dataset as most works did), but because we use Student's t, negative components, and SGHMC, there are some new parameters. Detailed parameter values can be found in our code. \textbf{The code is available at \href{https://github.com/realcrane/3D-student-splating-and-scooping}{https://github.com/realcrane/3D-student-splating-and-scooping.}}

All our experiments are running with one NVIDIA RTX 4090 GPU. We show training/rendering time in~\cref{tab:efficiency}. SSS is slower compared to vanilla 3DGS but still achieves real-time rendering (\textgreater70fps).

\begin{table}
\resizebox{0.48\textwidth}{!}{
\begin{tabular}{l|cc|cc|cc}
    Dataset & \multicolumn{2}{c|}{Mip-NeRF360}  & \multicolumn{2}{c|}{Tanks\&Temples} & \multicolumn{2}{c}{Deep Blending}\\
    Method|Metric & FPS & Training time & FPS & Training time& FPS & Training time \\
    \hline 
    3DGS & 99 & 21min & 120	& 12min	& 119 & 21min\\
    3DGS-MCMC & 79 & 32min & 105 & 17min & 138 & 29min\\
    SSS & 71 & 45min & 87 & 33min & 100 &21min\\
\end{tabular}
}
\caption{
    \textbf{Training time and rendering efficiency.} 
}
\label{tab:efficiency}
\end{table}

\subsection{Adaptation of Baselines}

To illustrate the parameter efficiency of our model, we did experiments with fewer components amount all baselines and our model. Because other baselines except 3DGS-MCMC do not support setting the maximum number of components, we modified their code. These modifications are minimal and do not involve any adjustment of hyperparameters to ensure a fair comparison. Specifically, we modified the adaptive density control used by most baselines. We stop densification (cloning and splitting) if the current number of components has reached the maximum setting. Note that due to the existence of the pruning strategy, the number of components can also be decreased, which will allow densification to continue until the preset densification iteration is reached.

\subsection{Ablation Study Implementation}
\label{sec:ablation_implementation}
We show our Ablation study results in \cref{sec:ablation_results}. These include three settings with one or more components in SSS to evaluate their effectiveness. Setting 1 (SGD + positive t-dis) uses adaptive density control and SGD optimization in vanilla 3DGS with positive Student's t distribution. Setting 2 (SGHMC + positive t-dis) uses SGHMC only with positive Student's t distribution. Setting 3 (full SSS model) uses SGHMC with both positive and negative Student's t distribution. Since Student's t distribution does not have an analytical form after convolution with another Student's t, we remove the low-pass filter in the splatting algorithm~\cite{zwicker2001ewa, zwicker2002ewa}. The main function of the low-pass filter is to ensure that the minimum scale of each component is close to 1 pixel. Removing the low-pass filter does not affect our final result, because our principled sampler can ensure that the final components are of the appropriate size. However, for setting 1, the absence of a low-pass filter will lead to an increase in the number of components (considering that it will use more tiny components to reconstruct more details). In order to ensure the fairness of the comparison (to make our number of components roughly equal to that of vanilla 3DGS), we added a small value (0.3) to the covariance matrix of the t-distribution after projection as a low-pass filter from an engineering perspective. Furthermore, we do not present the results of using adaptive density control and SGD optimization in vanilla 3DGS with both positive and negative t-distributions. This is because our negative components are designed to be used with our principled sampler. The densification of positive components in vanilla 3DGS is based on the size of the components and their gradients, which is not applicable to negative components. Adding negative components directly to adaptive density control will lead to worse results and is not the focus of our study. Finally, we also test our SGHMC sampler with Gaussian distributions to show its advantage. We use the same hyperparameters related to SGHMC and keep all other hyperparameters the same as vanilla 3DGS in this experiment.

\section{Forward and Backward Passes}
Although the general forward/backward passes of SSS are similar to 3DGS, the equations are different. This is mainly due to the introduction of t-distribution. Below we give details of mathematical derivation.

\subsection{Forward Pass}
\label{sec:forward_pass}
\paragraph{Affine transformation of 3D Student's t distribution}
A 3D Student's t distribution is:
\begin{align}
T(x&; \mu,\Sigma, \nu)=\frac{\Gamma(\frac{\nu+3}{2})}{(\nu\pi)^{\frac{3}{2}}\Gamma(\frac{\nu}{2})|\Sigma
|^{\frac{1}{2}}} \nonumber \\
&\cdot [1+\frac{1}{\nu}(x-\mu)^T\Sigma^{-1}(x-\mu)]^{-\frac{\nu+3}{2}}
\label{eq:tdis}
\end{align}
where $\nu\geq 1$, $x, \mu\in \mathbb{R}^3$ $\Sigma\in\mathbb{R}^{3\times 3}$ are the control parameter, the mean and the covariance matrix, which describes the spread and orientation of the distribution in 3D space. To render an image, \cref{eq:tdis} needs to be projected into a 2D image plane, which goes through a view transformation $W$, $d$, and a (approximate) projective transformation $J$~\cite{zwicker2001ewa, zwicker2002ewa}, so a 3D point $x$ after transformation becomes $u$:
\begin{align}
    u=m^{-1}(x)=A^{-1}(x-b) \nonumber\\
    \text{where } A=JW \text{ and } b=x+J(d-t)
    \label{eq:transform}
\end{align}
where $t$ is the camera coordinates. Applying the transformations \cref{eq:transform} to \cref{eq:tdis} gives:

\begin{equation}
\begin{aligned}
&T(x)=\frac{\Gamma(\frac{\nu+3}{2})}{(\nu\pi)^{\frac{3}{2}}\Gamma(\frac{\nu}{2})|\Sigma
|^{\frac{1}{2}}}\\
&\cdot [1+\frac{1}{\nu}(A^{-1}(x-b)-\mu)^T\Sigma^{-1}(A^{-1}(x-b)-\mu)]^{-\frac{\nu+3}{2}}\\
&=\frac{\Gamma(\frac{\nu+3}{2})}{(\nu\pi)^{\frac{3}{2}}\Gamma(\frac{\nu}{2})|\Sigma
|^{\frac{1}{2}}}\\
&\cdot [1+\frac{1}{\nu}(A^{-1}x-A^{-1}b-A^{-1}A\mu)^T\\
&\cdot \Sigma^{-1}(A^{-1}x-A^{-1}b-A^{-1}A\mu)]^{-\frac{\nu+3}{2}}\\
&=\frac{\Gamma(\frac{\nu+3}{2})}{(\nu\pi)^{\frac{3}{2}}\Gamma(\frac{\nu}{2})|\Sigma
|^{\frac{1}{2}}}\\
&\cdot [1+\frac{1}{\nu}(x-b-A\mu)^T(A^{-1})^T\Sigma^{-1}A^{-1}(x-b-A\mu)]^{-\frac{\nu+3}{2}}\\
&=\frac{\Gamma(\frac{\nu+3}{2})}{(\nu\pi)^{\frac{3}{2}}\Gamma(\frac{\nu}{2})|\Sigma
|^{\frac{1}{2}}}\\
&\cdot [1+\frac{1}{\nu}(x-m(\mu))^T(A\Sigma A^T)^{-1}(x-m(\mu))]^{-\frac{\nu+3}{2}}\\
&=\frac{\Gamma(\frac{\nu+3}{2})|A|}{(\nu\pi)^{\frac{3}{2}}\Gamma(\frac{\nu}{2})|\Sigma
|^{\frac{1}{2}}|A|^{\frac{1}{2}}|A^T|^{\frac{1}{2}}}\\
&\cdot [1+\frac{1}{\nu}(x-m(\mu))^T(A\Sigma A^T)^{-1}(x-m(\mu))]^{-\frac{\nu+3}{2}}\\
&=|A|\frac{\Gamma(\frac{\nu+3}{2})}{(\nu\pi)^{\frac{3}{2}}\Gamma(\frac{\nu}{2})|A\Sigma A^T|^{\frac{1}{2}}}\\
&\cdot [1+\frac{1}{\nu}(x-m(\mu))^T(A\Sigma A^T)^{-1}(x-m(\mu))]^{-\frac{\nu+3}{2}}\\
&=|A|T(x, m(\mu), A\Sigma A^T, \nu)\\
\end{aligned}
\end{equation}

So for $r(x)=T(x; \mu,\Sigma, \nu)$ at $x$, its transformed density is:
\begin{equation}
\begin{aligned}
&r_k^{'}(x)=\frac{1}{|A^{-1}|}T(x; m(\mu), A\Sigma A^T, \nu)\\
&=\frac{1}{|W^{-1}J_k^{-1}|}T(x; m(\mu), JW\Sigma W^TJ^T, \nu)\\
&=\frac{1}{|W^{-1}J^{-1}|}T(x; m(\mu),\Sigma^{'}, \nu)\\
\label{eq:transformed}
\end{aligned}
\end{equation}
where $\Sigma^{'}=JW\Sigma W^TJ^T$.

This result still holds after we drop the normalization constant $\frac{\Gamma(\frac{\nu+3}{2})}{(\nu\pi)^{\frac{3}{2}}\Gamma(\frac{\nu}{2})|\Sigma
|^{\frac{1}{2}}} \nonumber$.

\paragraph{Integral along a ray (marginalization)} After deriving the view and projective transformation for Student's t distribution, we need to integrate it along the ray that intersects with it for rendering. \ie splatting it. This is actually the marginalization of Student's t distribution along one dimension, and it can be done by another affine transformation $y=Wx$, where $W=\begin{bmatrix}\mathbf{I}, 0\\0, 0\\\end{bmatrix}$ with $\mathbf{I}$ is a $2\times 2$ identify matrix. To see this, imagine $x$ can be divided into two parts $x=\begin{bmatrix}x_1\\ x_2\end{bmatrix}$, so that:
\begin{equation}
\begin{aligned}
&T(Wx; \mu,\Sigma, \nu, p) = T(y; W\mu, W\Sigma W^T, \nu, p)\\
\end{aligned}
\end{equation}
and if $\mu= \begin{bmatrix} \mu_a\\\mu_b\\\end{bmatrix}$ and  
$\Sigma=
\begin{bmatrix}
    \Sigma_{aa} & \Sigma_{ab}\\
    \Sigma_{ba} & \Sigma_{bb}\\
\end{bmatrix}$ then 
\begin{equation}
\begin{aligned}
&T(x; \mu,\Sigma, \nu, p) = T(
\begin{pmatrix}
    x_a \\
    x_b
\end{pmatrix};
\begin{pmatrix}
    \mu_a \\
    \mu_b
\end{pmatrix},
\begin{pmatrix}
    \Sigma_{aa} & \Sigma_{ab} \\
    \Sigma_{ba} & \Sigma_{bb}
\end{pmatrix},
\nu,
p
)\\
\end{aligned}
\end{equation}
where means by transforming $x$ with $W=\begin{bmatrix}\mathbf{I}, 0\\0, 0\\\end{bmatrix}$, we will get the marginal distribution of $x_a$:
\begin{equation}
\begin{aligned}
T(Wx; \mu,\Sigma, \nu, p) &= T(y; W\mu, W\Sigma W^T, \nu, p)\\
&= T(x_a; \mu_a, \Sigma_{aa}, \nu, p_a)\\
\end{aligned}
\end{equation}
This way we can marginalize any variable for $T$. Integration along a ray gives a 2D t-distribution:
\begin{equation}
    T_{2D}(x, \mu, \Sigma, \nu)= [1+\frac{1}{\nu}(x-\mu)^T\Sigma^{-1}(x-\mu)]^{-\frac{\nu+2}{2}}
\end{equation}
where $x$ now is in 2D space.

So far, we have derived all key equations for the forward pass when using t-distributions as the mixture components.

\subsection{Backward Pass}
\label{sec:backward}
The most important computation in the backward pass is to compute all the key gradients, $\frac{\partial L}{\partial \mu}$, $\frac{\partial L}{\partial S}$ ,$\frac{\partial L}{\partial R}$, $\frac{\partial L}{\partial c}$, $\frac{\partial L}{\partial o}$, and $\frac{\partial L}{\partial \nu}$, where $L$ is the loss, $\mu$ is the mean of a t-distribution. $S$ and $R$ are the scaling and rotation matrices for the covariance matrix of t-distribution. $c$ is the color represented in spherical harmonics. $o$ is the opacity. $\nu$ is the control parameter of the t-distribution.

$L$ is calculated between the ground-truth pixel colors and rendered colors. The rendered pixel colors are:
\begin{equation}
    c(x) = \sum_{i=1}^{N}c_i o_i T^{2D}_{i}(x)\prod_{j=1}^{i-1}(1-o_jT^{2D}_{j}(x))
\end{equation}
where $x$ is the position of a pixel in 2D images.

For simplicity, we let $\alpha = oT$. Based on the chain rule:
\begin{equation}
    \frac{\partial_{L}}{\partial_{\mu}} = \frac{\partial_{L}}{\partial_{rgb}} \cdot \frac{\partial_{rgb}}{\partial_{\alpha}} \cdot \frac{\partial_{\alpha}}{\partial_{T^{2D}}} \cdot
    \frac{\partial_{T^{2D}}}{\partial_{h}}
    \cdot
    \frac{\partial_{h}}{\partial_{\mu'}}
    \cdot
    \frac{\partial_{\mu'}}{\partial_{\mu}}
\end{equation}
\begin{equation}
    \frac{\partial_{Loss}}{\partial_{o}} = \frac{\partial_{Loss}}{\partial_{rgb}} \cdot \frac{\partial_{rgb}}{\partial_{\alpha}} \cdot \frac{\partial_{\alpha}}{\partial_{o}} 
\end{equation}

\begin{equation}
    \frac{\partial_{L}}{\partial_{c}} = \frac{\partial_{L}}{\partial_{rgb}} \cdot \frac{\partial_{rgb}}{\partial_{c}}
\end{equation}
\begin{equation}
    \frac{\partial_{L}}{\partial_{S}} = \frac{\partial_{L}}{\partial_{rgb}} \cdot \frac{\partial_{rgb}}{\partial_{}} \cdot
    \frac{\partial_{\alpha}}{\partial_{T^{2D}}} \cdot
    \frac{\partial_{T^{2D}}}{\partial_{h}} \cdot
    \frac{\partial_{h}}{\partial_{\Sigma'}} \cdot
    \frac{\partial_{\Sigma'}}{\partial_{\Sigma}} \cdot \frac{\partial_{\Sigma}}{\partial_{S}}
\end{equation}
\begin{equation}
    \frac{\partial_{L}}{\partial_{R}} = \frac{\partial_{L}}{\partial_{rgb}} \cdot \frac{\partial_{rgb}}{\partial_{\alpha}} \cdot
    \frac{\partial_{\alpha}}{\partial_{T^{2D}}} \cdot
    \frac{\partial_{T^{2D}}}{\partial_{h}} \cdot
    \frac{\partial_{h}}{\partial_{\Sigma'}} \cdot
    \frac{\partial_{\Sigma'}}{\partial_{\Sigma}} \cdot
    \frac{\partial_{\Sigma}}{\partial_{R}}
\end{equation}
\begin{equation}
\label{eq:back_nu}
    \frac{\partial_{L}}{\partial_{\nu}} = \frac{\partial_{L}}{\partial_{rgb}} \cdot \frac{\partial_{rgb}}{\partial_{\alpha}} \cdot \frac{\partial_{\alpha}}{\partial_{T^{2D}}} \cdot
    \frac{\partial_{T^{2D}}}{\partial_{\nu}}
\end{equation}
where $\mu'$ and $\Sigma'$ are the projected $\mu$ and $\Sigma$ in 2D space. $h$ is a predefined function explained shortly.

For these gradients, we only need to replace the calculation of Gaussian in vanilla 3DGS with the calculation of Student's t distribution based on the chain rule, except for the gradient for $\nu$ in~\cref{eq:back_nu}. In order to further simplify the calculation, we extract the same part of the 2D Gaussian and 2D Student's t distributions, and define a function $h(x)$:
\begin{equation}
\begin{aligned}
    &G^{2D}(x, \mu', \Sigma') = exp^{-\frac{1}{2}h(x)}\\
    &T^{2D}(x, \mu', \Sigma', \nu)= [1+\frac{1}{\nu}h(x)]^{-\frac{\nu+2}{2}}\\
\end{aligned}
\end{equation}
where
\begin{equation}
\begin{aligned}
    &h(x) = (x-\mu')^T\Sigma'^{-1}(x-\mu')\\
\end{aligned}
\end{equation}

\noindent We have
\begin{equation}
\begin{aligned}
    &\frac{\partial T^{2D}(x, \mu', \Sigma', \nu)}{\partial h(x)} =[1 + \frac{1}{\nu}h(x)]^{-\frac{\nu+2}{2}} \quad dh(x)\\ 
    &= -\frac{\nu+2}{2} \cdot \frac{1}{\nu} \cdot [1+\frac{1}{\nu}h(x)]^{-\frac{\nu+4}{2}} \quad dh(x)\\
\end{aligned}
\end{equation}
to replace every
\begin{equation}
\begin{aligned}
     \frac{\partial G^{2D}(x, \mu', \Sigma')}{\partial h(x)} &=-\frac{1}{2}exp^{-\frac{1}{2}h(x)}\quad dh(x)\\
\end{aligned}
\end{equation}

Then we can use the derived calculations in 3DGS for the rest of the parts, we refer the readers to read~\cite{ye2023mathematical} for more detailed mathematics if interested.

Finally, the gradient for optimizing $\nu$ needs to be calculated separately. Assuming $g(\nu) = 1 + \frac{1}{\nu} h(x)$, we have
\begin{equation}
\begin{aligned}
    \frac{\partial_{T^{2D}(x, \mu', \Sigma', \nu)}}{\partial_{\nu}}&= [1+\frac{1}{\nu}(x-\mu')^T\Sigma'^{-1}(x-\mu')]^{-\frac{\nu+2}{2}} d\nu\\
    &=[1+\frac{1}{\nu}h(x)]^{-\frac{\nu+2}{2}} d\nu\\
    &=g(\nu)^{-\frac{\nu+2}{2}} d\nu\\
    &=-\frac{\nu+2}{2}g(\nu)^{-\frac{\nu+4}{2}} \cdot g^{'}(\nu)\\
    &=-\frac{\nu+2}{2}(1+\frac{1}{\nu}h(x))^{-\frac{\nu+4}{2}}\\ &\cdot ((1+\frac{1}{\nu}h(x)) d\nu)\\
    &=-\frac{\nu+2}{2}(1+\frac{1}{\nu}h(x))^{-\frac{\nu+4}{2}} \cdot (-\frac{h(x)}{\nu^2})\\
    &=\frac{\nu+2}{2} \cdot \frac{h(x)}{\nu^2} \left[1 + \frac{1}{\nu} h(x)\right]^{-\frac{\nu+4}{2}}\\
    &=\frac{\nu+2}{2} \cdot \frac{(x-\mu')^T\Sigma'^{-1}(x-\mu')}{\nu^2}\\
    &\cdot \left[1 + \frac{1}{\nu} (x-\mu')^T\Sigma'^{-1}(x-\mu')\right]^{-\frac{\nu+4}{2}}\\
\end{aligned}
\end{equation}

\section{Component Recycling}
\label{sec:relocation}
The key equation for relocating low opacity components to a high opacity component is to make sure the distribution before and after the relocation is not changed~\cite{rota2024revising, kheradmand20243d}. If we move new components to the location of an old component $\mu_{new} = \mu_{old}$, this is ensured by separately handling the opacity and the covariance matrix. For opacity, it is simply:
\begin{equation}
    (1 - O_{new})^N = (1-O_{old})
\end{equation}
For covariance:
\begin{equation}
\begin{aligned}
\label{eq:relocating}
    &minimize\int_{-\infty}^{\infty} ||C_{new}(x) - C_{old}(x) || dx \text{ or}\\
    &minimize||\int_{-\infty}^{\infty} C_{new}(x) - \int_{-\infty}^{\infty}C_{old}(x) || dx\\\
\end{aligned}
\end{equation}
To solve~\cref{eq:relocating}, we first need to separately derive $\int_{-\infty}^{\infty}C_{old}(x)$ and $\int_{-\infty}^{\infty}C_{new}(x)$. For $\int_{-\infty}^{\infty}C_{old}(x)$, assuming $u=\frac{x}{\sqrt{\nu_{old}\Sigma_{old}}}$, $du=\frac{1}{\sqrt{\nu_{old}\Sigma_{old}}}dx$, and $dx=\sqrt{\nu_{old}\Sigma_{old}}du$, then:
\begin{equation}
\begin{aligned}
\label{eq:integral_student_t}
    \int_{-\infty}^{\infty}C_{old}(x) = &\int_{-\infty}^{\infty} o_{old}[1+\frac{1}{\nu_{old}} \frac{x^2}{\Sigma_{old}}]^{-\frac{\nu_{old}+3}{2}} dx\\
    &= o_{old} \int_{-\infty}^{\infty} [1+\frac{1}{\nu_{old}} \frac{x^2}{\Sigma_{old}}]^{-\frac{\nu_{old}+3}{2}} dx\\
    &= o_{old} \int_{-\infty}^{\infty} [1+u^2]^{-\frac{\nu_{old}+3}{2}} \sqrt{\nu_{old}\Sigma_{old}}du\\
    &= o_{old} \sqrt{\nu_{old}\Sigma_{old}} \int_{-\infty}^{\infty} [1+u^2]^{-\frac{\nu_{old}+3}{2}}du\\
\end{aligned}
\end{equation}
For the form $\int_{-\infty}^{\infty} [1+x^2]^{-\alpha}dx$, assuming $x=tan(\theta)$, $dx=sec^2(\theta)d\theta$, and $1+x^2=1+tan^2(\theta)=sec^2(\theta)$, then:
\begin{equation}
\begin{aligned}
\label{eq:integral_beta}
    &\int_{-\infty}^{\infty} [1+x^2]^{-\alpha}dx\\
    &=2\int_{0}^{\infty} [1+x^2]^{-\alpha}dx\\
    &=2\int_{0}^{\pi/2} (sec^2(\theta))^{-\alpha} sec^2(\theta) d\theta\\
    &=2\int_{0}^{\pi/2} (sec^{2-2\alpha}(\theta))d\theta\\
    &=2\int_{0}^{\pi/2} (cos^{2\alpha-2}(\theta))d\theta\\
\end{aligned}
\end{equation}
Further, according to the definition of the $\beta$ function:
\begin{equation}
\begin{aligned}
\label{eq:beta}
    &\beta(x, y)=2\int_{0}^{\pi/2} sin^{2x-1}(\theta)cos^{2y-1}(\theta)d\theta\\
\end{aligned}
\end{equation}
\cref{eq:integral_beta} becomes:
\begin{equation}
\begin{aligned}
\label{eq:beta2}
    &2\int_{0}^{\pi/2} (cos^{2\alpha-2}(\theta))d\theta\\
    &2\int_{0}^{\pi/2} sin^{2\frac{1}{2} -1}(\theta)cos^{2(\alpha-\frac{1}{2})-1}(\theta)d\theta\\
    &=\beta(\frac{1}{2},(\alpha-\frac{1}{2}))\\
\end{aligned}
\end{equation}
Then \cref{eq:integral_student_t} becomes:
\begin{equation}
\begin{aligned}
\label{eq:integral_student_t2}
    &o_{old} \sqrt{\nu_{old}\Sigma_{old}} \int_{-\infty}^{\infty} [1+u^2]^{-\frac{\nu_{old}+3}{2}}du\\
    &=o_{old} \sqrt{\nu_{old}\Sigma_{old}}\cdot \beta(\frac{1}{2}, \frac{\nu_{old}+2}{2})\\
\end{aligned}
\end{equation}

\noindent For $\int_{-\infty}^{\infty} C_{new}(x)$,

\begin{equation}
\label{eq:integral_student_t_render}
\begin{aligned}
    &\int_{-\infty}^{\infty} C_{new}(x) = \int_{-\infty}^{\infty} \sum_{i=1}^N o_{new}[1+\frac{1}{\nu_{new}} \frac{x^2}{\Sigma_{new}}]^{-\frac{\nu_{new}+3}{2}}\\
    &\cdot (1-o_{new}[1+\frac{1}{\nu_{new}} \frac{x^2}{\Sigma_{new}}]^{-\frac{\nu_{new}+3}{2}})^{i-1} dx
\end{aligned}
\end{equation}

\noindent From Binomial theorem:
\begin{equation}
   (x+y)^n = \sum_{k=0}^{n}\begin{pmatrix}n\\k\end{pmatrix}x^{n-k}y^k=\sum_{k=0}^{n}\begin{pmatrix}n\\k\end{pmatrix}x^ky^{n-k} 
\end{equation}

\noindent \cref{eq:integral_student_t_render} becomes:
\begin{equation}
\begin{aligned}
\label{eq:integral_student_t_render2}
    &\int_{-\infty}^{\infty} \sum_{i=1}^N o_{new}[1+\frac{1}{\nu_{new}} \frac{x^2}{\Sigma_{new}}]^{-\frac{\nu_{new}+3}{2}}\\
    &\cdot (1-o_{new}[1+\frac{1}{\nu_{new}} \frac{x^2}{\Sigma_{new}}]^{-\frac{\nu_{new}+3}{2}})^{i-1} dx\\
    &=\int_{-\infty}^{\infty} \sum_{i=1}^N o_{new}[1+\frac{1}{\nu_{new}} \frac{x^2}{\Sigma_{new}}]^{-\frac{\nu_{new}+3}{2}}\\
    &\cdot \sum_{k=0}^{i-1}\begin{pmatrix}i-1\\k\end{pmatrix} (-o_{new}[1+\frac{1}{\nu_{new}} \frac{x^2}{\Sigma_{new}}]^{-\frac{\nu_{new}+3}{2}})^k dx\\
    &=\int_{-\infty}^{\infty} \sum_{i=1}^N o_{new}[1+\frac{1}{\nu_{new}} \frac{x^2}{\Sigma_{new}}]^{-\frac{\nu_{new}+3}{2}}\\
    &\cdot \sum_{k=0}^{i-1}\begin{pmatrix}i-1\\k\end{pmatrix} (-1)^k(o_{new})^k([1+\frac{1}{\nu_{new}} \frac{x^2}{\Sigma_{new}}]^{-\frac{\nu_{new}+3}{2}})^k dx\\
    &=\int_{-\infty}^{\infty} \sum_{i=1}^N \sum_{k=0}^{i-1}\begin{pmatrix}i-1\\k\end{pmatrix} (-1)^k(o_{new})^{k+1}\\
    &\cdot ([1+\frac{1}{\nu_{new}} \frac{x^2}{\Sigma_{new}}]^{-\frac{(k+1)(\nu_{new}+3)}{2}}) dx\\
    &=\sum_{i=1}^N \sum_{k=0}^{i-1}\begin{pmatrix}i-1\\k\end{pmatrix} (-1)^k(o_{new})^{k+1}\\
    &\cdot \int_{-\infty}^{\infty} ([1+\frac{1}{\nu_{new}} \frac{x^2}{\Sigma_{new}}]^{-\frac{(k+1)(\nu_{new}+3)}{2}}) dx\\
\end{aligned}
\end{equation}

\noindent According to equations~\ref{eq:integral_beta}, \ref{eq:beta} and \ref{eq:beta2}, \cref{eq:integral_student_t_render2} becomes:
\begin{equation}
\begin{aligned}
\label{integral_student_t_render3}
    &\sum_{i=1}^N \sum_{k=0}^{i-1}\begin{pmatrix}i-1\\k\end{pmatrix} (-1)^k(o_{new})^{k+1}\\
    &\cdot \int_{-\infty}^{\infty} ([1+\frac{1}{\nu_{new}} \frac{x^2}{\Sigma_{new}}]^{-\frac{(k+1)(\nu_{new}+3)}{2}}) dx\\
    &=\sum_{i=1}^N \sum_{k=0}^{i-1}\begin{pmatrix}i-1\\k\end{pmatrix} (-1)^k(o_{new})^{k+1}\\ 
    &\cdot \sqrt{\nu_{new}\Sigma_{new}}\cdot \beta(\frac{1}{2}, \frac{(k+1)(\nu_{new}+3)-1}{2})\\
\end{aligned}
\end{equation}

\noindent Having derived $\int_{-\infty}^{\infty}C_{old}(x)$ and $\int_{-\infty}^{\infty}C_{new}(x)$, we minimize \cref{eq:relocating} by setting:
\begin{equation}
\begin{aligned}
    &\int_{-\infty}^{\infty}C_{new}(x) = \int_{-\infty}^{\infty}C_{old}(x)\\
    \Rightarrow&\sum_{i=1}^N \sum_{k=0}^{i-1}\begin{pmatrix}i-1\\k\end{pmatrix} (-1)^k(o_{new})^{k+1}\\
    &\cdot \sqrt{\nu_{new}\Sigma_{new}} \beta(\frac{1}{2}, \frac{(k+1)(\nu_{new}+3)-1}{2})\\
    &= o_{old} \sqrt{\nu_{old}\Sigma_{old}} \beta(\frac{1}{2}, \frac{\nu_{old}+2}{2})\\
    \Rightarrow&\sqrt{\nu_{new}\Sigma_{new}} \sum_{i=1}^N \sum_{k=0}^{i-1}\begin{pmatrix}i-1\\k\end{pmatrix} (-1)^k(o_{new})^{k+1}\\
    &\cdot \beta(\frac{1}{2}, \frac{(k+1)(\nu_{new}+3)-1}{2})\\
    &= o_{old} \sqrt{\nu_{old}\Sigma_{old}} \beta(\frac{1}{2}, \frac{\nu_{old}+2}{2})\\
    \Rightarrow&\Sigma_{new} = (o_{old})^2 \frac{\nu_{old}}{\nu_{new}}\\
    &\resizebox{.47\textwidth}{!}{$(\frac{\beta(\frac{1}{2}, \frac{\nu_{old}+2}{2})}{\sum_{i=1}^N \sum_{k=0}^{i-1}\begin{pmatrix}i-1\\k\end{pmatrix} (-1)^k(o_{new})^{k+1} \beta(\frac{1}{2}, \frac{(k+1)(\nu_{new}+3)-1}{2})})^2$}\\
    &\cdot \Sigma_{old}\\
\end{aligned}
\end{equation}

\noindent So at the end, we can compute $\Sigma_{new}$ based on $\Sigma_{old}$:
\begin{equation}
\begin{aligned}
    &\Sigma_{new} = (o_{old})^2 \frac{\nu_{old}}{\nu_{new}}\\
    &\resizebox{.47\textwidth}{!}{$(\frac{\beta(\frac{1}{2}, \frac{\nu_{old}+2}{2})}{\sum_{i=1}^N \sum_{k=0}^{i-1}\begin{pmatrix}i-1\\k\end{pmatrix} (-1)^k(o_{new})^{k+1} \beta(\frac{1}{2}, \frac{(k+1)(\nu_{new}+3)-1}{2})})^2$}\\
    &\cdot \Sigma_{old}
\end{aligned}
\end{equation}

\noindent Furthermore, since $\beta$ function can be represented by $\Gamma$ functions:
\begin{equation}
\begin{aligned}
\label{eq:beta3}
    &\beta(x, y)=\frac{\Gamma(x)\Gamma(y)}{\Gamma(x+y)} \\
\end{aligned}
\end{equation}
We can use the $\Gamma$ function or $\ln(\Gamma)$ function instead of $\beta$ function in practice:
\begin{equation}
\begin{aligned}
\label{eq:beta4}
    \beta(x, y)&=\frac{\Gamma(x)\Gamma(y)}{\Gamma(x+y)} \\
    \Rightarrow\ln(\beta(x, y))&=\ln(\Gamma(x)) + \ln(\Gamma(y)) - \ln(\Gamma(x+y))\\
    \Rightarrow\beta(x, y)&=exp(\ln(\Gamma(x)) + \ln(\Gamma(y)) - \ln(\Gamma(x+y)))
\end{aligned}
\end{equation}

\section{SGHMC Sampling}
In SGHMC~\cite{chen2014stochastic}, the posterior distribution of model parameters $\theta$ given a set of independent observations $x\in D$ is defined as
\begin{equation}
    \pi(\theta, r) \propto exp(-U(\theta))
\end{equation}
where $U(\theta)$ is a potential energy function which is $-\sum_{x\in D}\log p(x|\theta)-\log p(\theta)$.

To sample from $p(\theta|D)$, The Hamiltonian (Hybrid) Monte Carlo (HMC) considers generating samples from a joint distribution of $\pi(\theta, r)$ defined by
\begin{equation}
    \pi(\theta, r) \propto exp(-U(\theta) -\frac{1}{2}r^TMr)
    \label{eq:posterior_supp}
\end{equation}
where the Hamiltonian function is defined by $H(\theta, r)=U(\theta)+\frac{1}{2}r^TMr$. $M$ is the mass matrix and $r$ is the auxiliary momentum variables. Further, to introduce stochastic gradients into the sampling, the Stochastic Gradient Hamiltonian Monte Carlo (SGHMC) is proposed in \cite{chen2014stochastic} where an additional friction is introduced. We refer the readers to \cite{chen2014stochastic} for detailed mathematical derivation.

To employ SGHMC for our SSS sampling, we start by parameterizing a joint distribution:
\begin{equation}
    P(\theta, r) \propto exp(-L_{\theta}(x) -\frac{1}{2}r^TIr)
    \label{eq:posterior1}
\end{equation}
By defining a similar Hamiltonian function as in \cref{eq:posterior_supp} and adding a friction term as in \cite{chen2014stochastic}, we derive
\begin{align}
    d\theta = &M^{-1}rdt \nonumber \\
    dr = &-\nabla U(\theta)dt - CM^{-1}rdt + \mathcal{N}(0, 2Cdt)
\label{eq:update_supp}
\end{align}

Next, we further modify the updating equations in \cref{eq:update_supp} to:

\begin{align}
    \mu_{t+1} &= \mu_t - \varepsilon^2 \left [\frac{\partial L}{\partial \mu}\right ]_{t} + F + N \nonumber \\
    F &= \sigma(o)\varepsilon(1 - \varepsilon C)r_{t-1} \nonumber \\
    N &= \sigma(o)\mathcal{N}(0, 2\varepsilon^{\frac{3}{2}}C) \nonumber \\
    r_{t+1} &= r_{t} - \varepsilon\left[\frac{\partial L}{\partial \mu}\right]_{t+1} - \varepsilon C r_{t-1} + \mathcal{N}(0, 2\varepsilon C) \nonumber \\
    &\text{where }\sigma(o) = \sigma(-k(o-t))
    \label{eq:sigmoidUpdate_supp}
\end{align}
where $\varepsilon$ is the learning rate and decays during learning. $\mathcal{N}$ is Gaussian noise. $o$ is the opacity. To further clarify the relation between the learnable parameter and the momentum, we first show the updating rule for the learnable parameter in the original SGHMC:
\begin{equation}
\footnotesize
\begin{aligned}
    \mu_{t+1} &= \mu_t + \varepsilon*(r_t - \varepsilon* G_t - \varepsilon * C * r_t + N(0, 2*\varepsilon*C))\\
    &= \mu_t + \varepsilon*r_t - \varepsilon^2 * G_t - \varepsilon^2 * C * r_t + N(0, 2 * \varepsilon^{\frac{3}{2}} *C)\\
    &= \mu_t - \varepsilon^2 * G_t + \varepsilon * (1 - \varepsilon * C) * r_t + N(0, 2 * \varepsilon^{\frac{3}{2}} *C)\\
\label{eq:SGHMC}
\end{aligned}
\end{equation}
where $G$ is the gradient $\left [\frac{\partial L}{\partial \mu}\right ]$. 

Before Stochastic Gradient Hamiltonian Monte Carlo (SGHMC), we first attempted the Stochastic Gradient Langevin Dynamics (SGLD) sampling in 3DGS-MCMC~\cite{kheradmand20243d}. Although it outperforms the standard optimization employed in the original 3DGS and its variants, it still sometimes generates suboptimal results. Other than the randomness in the optimization itself, we suspected the core reason is the increased model complexity, especially the introduction of $\nu$ in t-distribution, which brings tight coupling between many parameters, \eg $\nu$ greatly influencing $\mu$ and $\Sigma$. On the high level, the optimization of $\mu$ and $\Sigma$ can be seen as seeking the optimal distribution within a family of distributions. In 3DGS, this family is Gaussians. However, when $\nu$ is also optimized, the family itself changes during optimization. This is the core reason we resort to SGHMC which shows better sampling behaviors given tightly coupled parameters~\cite{chen2014stochastic}.

Furthermore, we leave the learning of $\nu$ and other parameters to Adam, as this can help further decouple the parameters. This is a similar strategy to the 3DGS-MCMC. This is complemented by using SGHMC on the location of t-distribution $\mu$. Also, for components with high opacity, we tend to think that they are near their local optima, so no further random perturbation is needed. This is achieved by adding a sigmoid switch:
\begin{equation}
\scriptsize
\begin{aligned}
    \mu_{t+1} = \mu_t - \varepsilon^2 * G_t + \sigma(\varepsilon * (1 - \varepsilon * C) * r_t) + \sigma(N(0, 2 * \varepsilon^{\frac{3}{2}} *C))\\
\end{aligned}
\end{equation}
$\sigma$ is the customized sigmoid function. Note we add the sigmoid switch to both the friction and the noise, partially to keep the integrity of the sampler and partially to remove the friction for nearly optimal components. Also, when the friction and noise are removed, the parameter is updated by $\varepsilon^2 * G_t$, \ie the gradient scaled by $\varepsilon^2$ which is much smaller than the learning rate $\varepsilon$, encouraging local search.

Finally, we conducted experiments between our SGHMC and SGLD in 3DGS-MCMC. We found that while SGLD can explore large spaces, SGHMC is better at local exploitation. To achieve the best results, we finally performed a burn-in stage with the friction removed during training for large exploration. In order to maintain the anisotropy of $\Sigma$ of Student's t distribution after the friction is removed, we multiply the noise by $\Sigma$ following 3DGS-MCMC. After the burn-in stage, we add the friction back and restore the noise (no longer multiplied by $\Sigma$).

\section{Representation Limitation}
Although we demonstrate the strength of our approach in both qualitative and quantitative evaluations, we do acknowledge that our approach is not perfect in every scenario. We have discussed the limitations of SSS in the main context. For example, the Student's t distribution is limited by symmetric and smooth representation, which makes it difficult to handle sharp shapes perfectly. Student's t distribution combined with negative components can increase the representation ability, but the range of representation is still limited. In addition, although the randomness of SGHMC brings more exploration of space, it still sometimes suffer from the floating artifact problem commonly observed in 3DGS.


\end{document}